\documentclass[10pt,final,journal,twocolumn,singlespaced]{IEEEtran}
\usepackage{algorithm}
\usepackage{algpseudocode}
\usepackage{amsmath}
\usepackage{amssymb}
\usepackage{bm}
\usepackage{booktabs}
\usepackage{color}
\usepackage{cite}
\usepackage{epsfig}
\usepackage{epstopdf}
\usepackage{flafter}
\usepackage{float}
\usepackage{ifthen}
\usepackage{multirow}
\usepackage{makecell} 
\usepackage{pifont}
\usepackage{subfigure}
\usepackage{times}
\usepackage{threeparttable}
\usepackage{url}  
\usepackage{subfigure,graphicx}
\usepackage{float}
\usepackage{color,multirow}
\usepackage{makecell}
\usepackage{cite} 
\usepackage{url}  
\usepackage{ifthen}  
\usepackage[T1]{fontenc}
\usepackage{graphicx}
\usepackage{flushend}
\usepackage[colorlinks,linkcolor=red,anchorcolor=gray,citecolor=green,urlcolor=black]{hyperref}
\usepackage{graphics}
\usepackage{booktabs}

\newcommand{\argmin}{\mathop{\mathrm{argmin}}}

\newcommand{\tabincell}[2]{\begin{tabular}{@{}#1@{}}#2\end{tabular}}

\begin{document}
\title{Nonlocal Patch-Based Fully-Connected Tensor Network Decomposition for Remote Sensing Image Inpainting
\thanks{This work was supported in part by the National Natural Science Foundation of China under Grant 61876203 and Grant 12071345, in part by the Project for Applied Basic Research of Sichuan Province under Grant 2021YJ0107, and in part by the  Program for Science and Technology Development of Henan Province under Grant 212102210511. \emph{(Corresponding authors: Xi-Le Zhao and Yu-Bang Zheng.)}}
\thanks{W.-J. Zheng, X.-L. Zhao, and Y.-B. Zheng are with the School of Mathematical Sciences, University of Electronic Science and Technology of China, Chengdu 611731, China (e-mail: wjz1355@163.com; xlzhao122003@163.com; zhengyubang@163.com).}
\thanks{Z.-F. Pang is with the School of Mathematics and Statistics, Henan University, Kaifeng 475004, China (e-mail: zhifengpang@163.com).}
}

\author{Wen-Jie Zheng, 
Xi-Le Zhao,
Yu-Bang Zheng,
and Zhi-Feng Pang
}
\maketitle

\begin{abstract}
Remote sensing image (RSI) inpainting plays an important role in real applications. Recently, fully-connected tensor network (FCTN) decomposition has been shown the remarkable ability to fully characterize the global correlation. Considering the global correlation and the nonlocal self-similarity (NSS) of RSIs, this paper introduces the FCTN decomposition to the whole RSI and its NSS groups, and proposes a novel nonlocal patch-based FCTN (NL-FCTN) decomposition for RSI inpainting. Different from other nonlocal patch-based methods, the NL-FCTN decomposition-based method, which increases tensor order by stacking similar small-sized patches to NSS groups, cleverly leverages the remarkable ability of FCTN decomposition to deal with higher-order tensors. Besides, we propose an efficient proximal alternating minimization-based algorithm to solve the proposed NL-FCTN decomposition-based model with a theoretical convergence guarantee. Extensive experiments on RSIs demonstrate that the proposed method achieves the state-of-the-art inpainting performance in all compared methods.
\end{abstract}

\section{Introduction}\label{secInt}

\begin{figure*}[!t]
	\footnotesize
	\setlength{\tabcolsep}{10pt}
	\begin{center}
		\begin{tabular}{c}
			\includegraphics[width=0.95\textwidth]{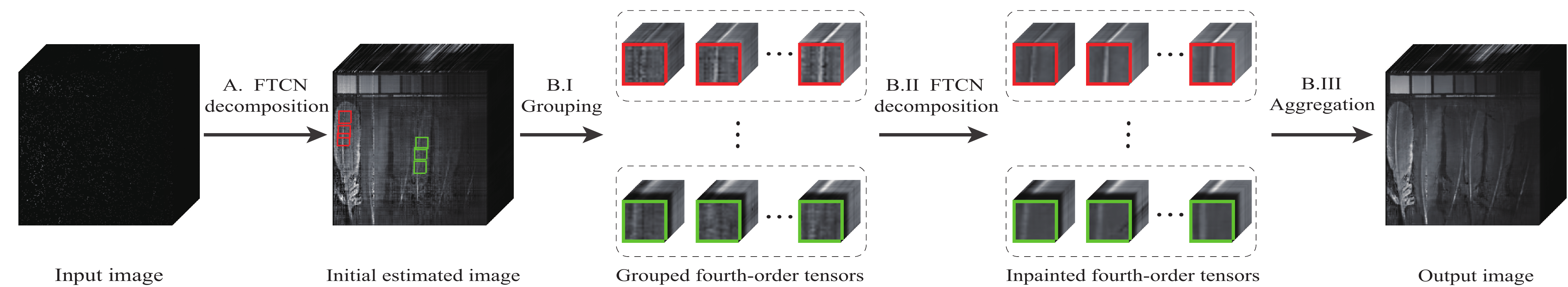}
		\end{tabular}
	\end{center}
	\caption{Flowchart of the proposed RSI inpainting method. It involves two stages: A. initial FCTN decomposition inpainting and B. nonlocal FCTN decomposition inpainting. B includes three steps consisted of grouping, grouped FCTN decomposition inpainting, and aggregation.}
	\label{illu_NLFCTN}
\end{figure*}

\IEEEPARstart{R}{emote} sensing images (RSIs) contain a wealth of spatial, spectral, and temporal information, which makes them widely applied in various applications\cite{sun2019lateral,hong2020learning,hong2021interpretable}. In practice, however, RSIs obtained are often incomplete due to some unavoidable factors, such as dead pixels and cloud cover. In order to improve the subsequent practicability, it is important to estimate missing elements by RSI inpainting\cite{wang2021total,yuan2018hyperspectral,chang2013simultaneous}. The core problem of RSI inpainting is to characterize the intrinsic prior of RSIs\cite{he2019non,zhang2019hybrid,zhuang2021hyperspectral,zhuang2020hyperspectral}.

Global correlation is an important prior of RSIs. Since RSIs are usually modeled as higher-order tensors, whose global correlation can be captured by tensor decomposition. There are various types of tensor decomposition, such as CANDECOMP/PARAFAC (CP) decomposition, Tucker decomposition, tensor singular value decomposition (t-SVD), tensor train (TT) decomposition, and tensor ring (TR) decomposition. All of them have been increasingly applied to RSI restoration and achieved promising results. For example, Liu et al.\cite{liu2012denoising} proposed a CP decomposition-based RSI denoising model, and Ng et al.\cite{ng2017adaptive} introduced a Tucker decomposition-based adaptive weighted sum of the nuclear norm (SNN) for RSI inpainting. Based on t-SVD, Fan et al.\cite{fan2017hyperspectral} employed tensor nuclear norm (TNN) to RSI restoration, and Zheng et al.\cite{zheng2019mixed} further expanded TNN to convex three-directional tensor nuclear norm and non-convex three-directional log-based tensor nuclear norm, and proposed two models for mixed noise removal. TT and TR decompositions as two representative tensor network decompositions received extensive applications in image restoration, since they have the superior capability to deal with higher-order tensors. For example, Bengua et al.\cite{bengua2017efficient} first proposed a TT decomposition-based image inpainting model, and Ding et al.\cite{ding2019low} further embedded a total variation (TV) regularization into TT decomposition to improve the capability of image details recovery. Based on TR decomposition, Yuan et al.\cite{yuan2019tensor} and He et al.\cite{he2019remote} proposed an image inpainting method by combining factor regularization and TV regularization, respectively. 

However, the methods directly considered the global correlation, which ignored the redundancy of repeated local patches across the spatial modes of RSIs, i.e., nonlocal self-similarity (NSS). One way to fully utilize NSS prior is to apply tensor low-rank method, such as log-based SNN \cite{xie2018tensor} and TNN, to each NSS group \cite{song2018nonlocal}, which is a new tensor obtained by stacking similar small-sized patches. By taking a third-order multispectral image (MSI) as an example, each original NSS group is a fourth-order tensor, including two spatial modes, one spectral mode, and one similar-group mode. However, traditional methods usually reshaped two spatial modes in each NSS group to one mode, which cannot better preserve the spatial structure of each NSS group. Therefore, Ding et al.\cite{ding2021tensor} exploited TT rank minimization to the original fourth-order NSS groups and achieved promising completion results. But TT decomposition only characterizes the correlation between adjacent two modes, causing a limited characterization for the NSS groups. 

Recently, a fully-connected tensor network (FCTN) decomposition has been proposed, which establishes an operation between any two modes and has the powerful capability to capture the global correlation of tensors\cite{zheng2021fully}. Based on FCTN decomposition, this paper makes two-fold contributions listing as follows:

\begin{table}[!t]
	\footnotesize
	\setlength{\tabcolsep}{1.5pt}
	\renewcommand\arraystretch{1.2}
	\caption{Notation declarations.}
	\begin{center}
		\begin{tabular}{c|c}
			\Xhline{0.8pt}
			Notations              & Interpretations  \\
			
			\hline\hline
			
			\tabincell{c}{$x$, $\mathbf{x}$, $\mathbf{X}$, $\mathcal{X}$}      
			&   scalar, vector, matrix, tensor \\
			\tabincell{c}{$\mathcal{X}({i_1, i_{2}, \cdots , i_N})$}   &  the $(i_1, i_{2}, \cdots , i_N)$th element of $\mathcal{X}$ \\
			\tabincell{c}{$\mathcal{X}_{1:d}$}
			& $(\mathcal{X}_1, \mathcal{X}_2, \cdots, \mathcal{X}_d)$ \\
			\tabincell{c}{$\|\mathcal{X}\|_F$} 
			& \tabincell{c}{$\|\mathcal{X}\|_F=\sqrt{\sum_{i_1, i_2, \cdots , i_N}|\mathcal{X}({i_1, i_{2}, \cdots , i_N})|^2}$}\\            
			\Xhline{0.8pt}
		\end{tabular}
	\end{center}
	\label{Nota_dec}
\end{table}

First, we suggest a novel nonlocal patch-based FCTN (NL-FCTN) decomposition-based RSI inpainting model, the flowchart of which is illustrated in Fig \ref{illu_NLFCTN}. As shown in Fig \ref{illu_NLFCTN}, the proposed model has two advantages. On the one hand, employing FCTN decomposition to the whole RSI and its NSS groups fully develops the global correlation and the NSS of RSIs. On the other hand, NL-FCTN decomposition-based method regards the process of stacking similar small-sized patches to NSS groups as a tensor order increment operation, which cleverly leverages the remarkable ability of FCTN decomposition on dealing with higher-order tensors.  

Second, we develop a proximal alternating minimization (PAM)-based algorithm to solve the NL-FCTN decomposition-based model with a theoretical guarantee of
convergence. Extensive experiments show that the proposed method outperforms the state-of-the-art methods on RSI inpainting.

\section{Notations and Preliminaries}\label{sec_nota}

We summarize the notations throughout this paper in Table \ref{Nota_dec}, and introduce a generalized tensor unfolding operation following \cite{zheng2021fully}. Supposing that $\mathbf{n}$ is a reordering of the vector $(1,2, \dots, N)$, the generalized tensor unfolding represents an $N$th-order tensor  $\mathcal{X}\in\mathbb{R}^{I_1\times I_2\times\cdots\times I_N}$ as a matrix $\mathbf{X}\in\mathbb{R}^{\prod_{i=1}^{d}I_{n_i}\times \prod_{i=d+1}^{N}I_{n_i}}$, which is denoted by $\mathbf{X}_{[n_{1:d};n_{d+1:N}]}={\tt GenUnfold}(\mathcal{X},n_{1:d};n_{d+1:N})$. On the contrary, its inverse operation is denoted by $\mathcal{X}={\tt GenFold}(\mathbf{X}_{[n_{1:d};n_{d+1:N}]}, n_{1:d}; n_{d+1:N})$.

\section{Proposed RSI Inpainting Method}\label{sec_model}

In this section, we start with a brief introduction to the definition and properties of FCTN decomposition, and then introduce the proposed NL-FCTN decomposition-based method for RSI inpainting in detail.

\subsection{FCTN Decomposition}

Compared with TT and TR decompositions, FCTN decomposition is more efficient and more genreal. Specifically, by factorizing an $N$th-order tensor into a sequence of small-sized $N$th-order factor tensors and establishing the relationship between any two factors, FCTN decomposition has the remarkable ability to fully characterize the global correlation of tensors. Mathematically, given an $N$th-order tensor $\mathcal{X}\in \mathbb{R}^{I_1\times I_2\times\cdots\times I_N}$, FCTN decomposition denotes each element of $\mathcal{X}$ by
\begin{equation}
	\begin{aligned}
		\mathcal{X}&(i_1,i_2,\cdots,i_N)=\\
		&\!\!\!\!\sum_{r_{1,2}=1}^{R_{1,2}}\sum_{r_{1,3}=1}^{R_{1,3}}\!\!\cdots\!\!\sum_{r_{1,N}=1}^{R_{1,N}}
		\sum_{r_{2,3}=1}^{R_{2,3}}\!\!\cdots\!\!\sum_{r_{2,N}=1}^{R_{2,N}}\!\!\cdots\!\!\sum_{r_{N-1,N}=1}^{R_{N-1,N}}\\
		\big\{&\mathcal{G}_1(i_1,r_{1,2},r_{1,3},\!\cdots\!,r_{1,N})\\
		&\mathcal{G}_2(r_{1,2},i_2,r_{2,3},\!\cdots\!,r_{2,N})\!\cdots\!\\
		&\mathcal{G}_k(r_{1,k},r_{2,k},\!\cdots\!,r_{k-1,k},i_k,r_{k,k+1},\!\cdots\!,r_{k,N})\!\cdots\!\\
		&\mathcal{G}_N(r_{1,N},r_{2,N},\!\cdots\!,r_{N-1,N},i_N)\big\},
	\end{aligned} \label{TNelement}
\end{equation}
where $\mathcal{G}_k\in\mathbb{R}^{R_{1,k}\times R_{2,k}\times\cdots\times R_{k-1,k}\times I_k\times R_{k,k+1}\times\cdots\times R_{k,N}}~(k=1,2,\cdots,N)$ are called the FCTN-factors and the vector $(R_{1,2},R_{1,3},\!\cdots\!,R_{1,N},R_{2,3},\!\cdots\!,R_{2,N},\!\cdots\!,R_{N-1,N})$ is called the FCTN-rank. For the sake of simplicity, the equation (\ref{TNelement}) can be expressed as $\mathcal{X}=\text{FCTN}(\mathcal{G}_1,\mathcal{G}_2,\cdots,\mathcal{G}_N)$.

\begin{figure}[!t]
	\footnotesize
	\setlength{\tabcolsep}{4pt}
	\begin{center}
		\begin{tabular}{c}
			\includegraphics[width=0.44\textwidth]{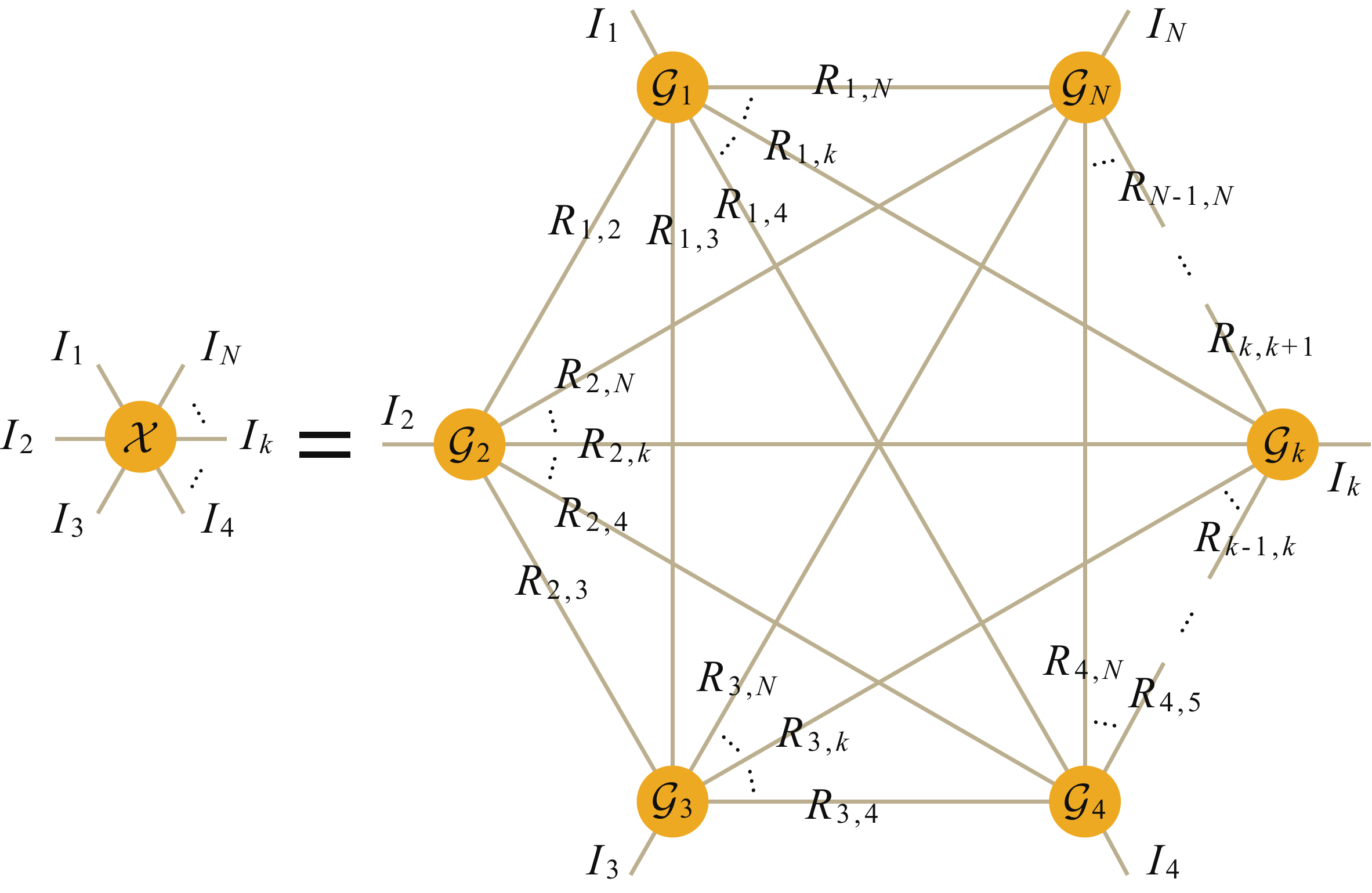}\\
		\end{tabular}
	\end{center}
	\caption{Illustration of FCTN decomposition.}
	\label{illu_FCTN}
\end{figure}

Fig. \ref{illu_FCTN} vividly illustrates the graphic structure of FCTN decomposition. In Fig. \ref{illu_FCTN}, we can clearly observe that any two FCTN factors are linked by a line, representing a multi-linear operation between them. 

\subsection{NL-FCTN Decomposition-Based RSI Inpainting Method}

The following is a detailed derivation of the proposed NL-FCTN decomposition-based model for RSI inpainting with the Fig.\ref{illu_NLFCTN}. The model consists of two stages. In stage A, we introduce FCTN decomposition to the whole RSI as an initialization process, guaranteeing the high accuracy of the subsequent group matching.

In stage B, there are three steps. In step B.I, we divide the initial RSI into small-sized patches, and stack similar patches as NSS groups. By taking an MSI $\mathcal{F}\in\mathbb{R}^{I_1\times I_2\times I_3}$ as an example, we divide it into overlapped cubes $\{\mathcal{F}_t\in\mathbb{R}^{p\times p\times I_3}\}_{t=1}^{T}$ with patch size $p$, where $T=((I_1-p)/(p-o)+1)\times((I_2-p)/(p-o)+1)$ is the number of total patches and $o$ is the size of overlap. Then, we select key patches from total patches with a fixed interval of $v$. $L=((I_1-p)/v+1)\times((I_2-p)/v+1)$ is the number of key patches as well as the number of NSS groups. This is because an NSS group consists of cubes stacking by a key patch and its similar patches. For each key patch, its similar patches are obtained by block matching. The way is to utilize the Euclidean distance to access similarity between each key patch and all total patches, and cluster the first $s$ similar patches into an NSS group. The NSS groups of MSI are denoted by $\{\hat{\mathcal{F}}_l\in\mathbb{R}^{p\times p\times I_3\times s}\}_{l=1}^L.$ Thus, a third-order MSI $\mathcal{F}$ transforms into a set of fourth-order NSS groups $\{\hat{\mathcal{F}}_l\}_{l=1}^L$ by group matching. Following a similar step described above, a fourth-order time-series RSI with the size of $I_1\times I_2\times I_3\times I_4$ transforms into a set of fifth-order NSS groups with the size of $p\times p\times I_3\times I_4\times s$. More generally, an $N$th-order tensor naturally transforms into a set of $(N+1)$th-order NSS groups.

\begin{algorithm}[!t]
	\small
	\caption{\small{PAM-Based Algorithm for NL-FCTN Decomposition-Based RSI Inpainting Model}}
	\begin{algorithmic}[1]
		\renewcommand{\algorithmicrequire}{\textbf{Input:}}
		\Require
		The degraded RSI $\mathcal{T} \in \mathbb{R}^{I_1\times I_2\times\cdots\times I_N}$, the index set $\Omega$, the patch size $p$, the similar patch number $s$, and the FCTN-rank $R$. 
		\renewcommand{\algorithmicrequire}{\textbf{Initialization:}}
		\Require
		The initial iterations $q=0$, the maximum iterations $q_{\text{max}}=100$, the overlapped patch size $o=1$, the key patch interval $v=p-1$, and $\rho=0.01$.
		\State{\textbf{Stage A. initial FCTN decomposition inpainting}}
		\State Inpaint the degraded RSI $\mathcal{T}$ via FCTN decomposition and obtain an initial RSI $\mathcal{F}$.
		\State{\textbf{Stage B. nonlocal FCTN decomposition inpainting}}
		\State (I) Group similar patches over all bands of $\mathcal{F}$ and obtain a set of $(N+1)$th-order NSS groups $\{\hat{\mathcal{F}}_l\}_{l=1}^L$.
		\State (II) Inpaint each NSS group $\hat{\mathcal{F}}_l$ via FCTN decomposition
		\For {$l=1:L$}\vspace{0.05cm}
		\State Update $\hat{\mathcal{F}}_l$ via (\ref{solve_group}) and obtain $\hat{\mathcal{X}}_l$.
		\EndFor
		\State (III) Aggregate the inpainted NSS groups $\{\hat{\mathcal{X}}_l\}_{l=1}^{L}$ to form the inpainted RSI $\mathcal{X}$.
		\Ensure
		The inpainted RSI $\mathcal{X}$.
	\end{algorithmic}
	\label{algorithms_NLFCTN}
\end{algorithm}

In step B.II, we regard an NSS group as a basis inpainting unit beacuse of its strong global correlation, and introduce FCTN decomposition to each NSS group. For FCTN decomposition, this step not only makes full use of the remarkable ability to characterize the global correlation, but also cleverly leverages the advantage of dealing with higher-order tensors by NSS-based tensor order increment operation. 

Without loss of generality, we consider an $(N+1)$th-order NSS group $\hat{\mathcal{F}}_l$ and employ FCTN decomposition to obtain the inpainted NSS group $\hat{\mathcal{X}}_l$, which can be formulated as
\begin{equation}
	\begin{aligned}
		\min_{\hat{\mathcal{X}}_l,\mathcal{G}}&~~\frac{1}{2}\big\|\hat{\mathcal{X}}_l-\text{FCTN}(\mathcal{G}_1,\mathcal{G}_2,\cdots,\mathcal{G}_{N+1})\big\|_F^2\\
		\text{s.t.}&~~\mathcal{P}_{\Omega_l}(\hat{\mathcal{X}}_l)=\mathcal{P}_{\Omega_l}(\hat{\mathcal{F}}_l),
	\end{aligned} \label{group_model}
\end{equation} 
where $\mathcal{P}_{\Omega_l}(\cdot)$  is the projection operator that keeps the entries in $\Omega_l$ and sets others to zero, and $\Omega_l$ is the index of known elements in $\hat{\mathcal{F}}_l$. Since (\ref{group_model}) is convex for $\hat{\mathcal{X}}_l$ and the FCTN-factors $\mathcal{G}_i (i=1,2,\cdots,N+1)$ independently, PAM \cite{attouch2013convergence} is introduced to obtain the solution as follows:
\begin{equation}
	\left\{
	\begin{aligned}
		&\!\!\mathcal{G}_i^{(q+1)}\!\!=\!\argmin_{\mathcal{G}_i}\!\Big\{\!\frac{1}{2}\big\|\hat{\mathcal{X}}_l^{(q)}\!\!-\text{FCTN}(\mathcal{G}_{1:i-1}^{(q+1)},\mathcal{G}_i,\mathcal{G}_{i+1:N+1}^{(q)})\big\|_F^2\\
		&~~~~~~~~~~~~~~~~~~~~+\!\frac{\rho}{2}\big\|\mathcal{G}_i\!-\!\mathcal{G}_i^{(q)}\big\|_F^2\!\Big\},~i\!=\!1,2,\cdots,N+1,\\
		&\!\!\hat{\mathcal{X}}_l^{(q+1)}\!\!=\!\!\!\!\argmin_{\mathcal{P}_{\Omega_l}(\hat{\mathcal{X}}_l)=\mathcal{P}_{\Omega_l}(\hat{\mathcal{F}}_l)}\!\!\!\Big\{\!\frac{1}{2}\!\big\|\hat{\mathcal{X}}_l\!-\!\text{FCTN}(\mathcal{G}_{1:N+1}^{(q+1)}\!)\big\|_F^2\!\!\\
		&~~~~~~~~~~~~~~~~~~~~~~~~~~~+\!\frac{\rho}{2}\!\big\|\hat{\mathcal{X}}_l\!-\!\hat{\mathcal{X}}_l^{(q)}\!\big\|_F^2\!\Big\},
	\end{aligned}
	\right.
	\label{group_solve}
\end{equation}
where $\mathcal{G}_i^{(q)} (i=1,2,\cdots,N+1)$ and $\hat{\mathcal{X}}_l^{(q)}$ are the results of the $q$th iteration of $\mathcal{G}_i (i=1,2,\cdots,N+1)$ and $\hat{\mathcal{X}}_l$ respectively, and $\rho>0$ is a proximal parameter. According to Theorem 4 in \cite{zheng2021fully}, 
\begin{equation*}
	\hat{\mathbf{X}}_{l(i)}=(\mathbf{G}_i)_{(i)}(\mathbf{M}_i)_{[m_{1:N};n_{1:N}]},
\end{equation*}
where $\mathcal{M}_i=\text{FCTN}\big(\{\mathcal{G}_i\}_{i=1}^{N+1},/\mathcal{G}_i\big)$ is a composition of $\mathcal{G}_1,\mathcal{G}_2,\cdots,\mathcal{G}_{i-1},\mathcal{G}_{i+1},\cdots,\mathcal{G}_{N+1},$
\begin{equation*}
	m_j=\left\{
	\begin{aligned}
		&2j,~~~~~~\text{if}~j<i,\\
		&2j-1,~\text{if}~j\geq i,
	\end{aligned}
	\right.
	~~\text{and}~~n_j=\left\{
	\begin{aligned}
		&2j-1,~\text{if}~j<i,\\
		&2j,~~~~~~\text{if}~j\geq i.
	\end{aligned}
	\right.
\end{equation*}
Thus, the $\mathcal{G}_i(i=1,2,\cdots,N+1)$-subproblems are given by
\begin{equation}
	\begin{aligned}
		\mathcal{G}_i^{(q+1)}\!&=\!\argmin_{\mathcal{G}_i}\Big\{\!\frac{\rho}{2}\big\|\mathcal{G}_i-\mathcal{G}_i^{(q)}\big\|_F^2\\
		&+\!\frac{1}{2}\big\|\hat{\mathbf{X}}^{(q)}_{l(i)}\!-\!(\mathbf{G}_i)_{(i)}(\mathbf{M}_i^{(q)})_{[m_{1:N};n_{1:N}]}\big\|_F^2\!\Big\}.
		\label{solveG}
	\end{aligned}
\end{equation}
It is easy to find that (\ref{group_solve}) and (\ref{solveG}) have the closed-form solutions as
\begin{equation}
	\left\{
	\begin{aligned}
		&\!\!\mathcal{G}_i^{(q+1)}\!\!=\!{\tt GenFold}\Big(\!\big[\hat{\mathbf{X}}^{(q)}_{l(i)}(\mathbf{M}_i^{(q)})_{[n_{1:N};m_{1:N}]}\!+\!\rho(\mathbf{G}_i^{(q)})_{(i)}\big]\\
		&~~~~~~~~~~\big[(\mathbf{M}_i^{(q)})_{[m_{1:N};n_{1:N}]}(\mathbf{M}_i^{(q)})_{[n_{1:N};m_{1:N}]}\!+\!\rho\mathbf{I}\big]^{-1}\!\Big),\\
		&~~~~~~~~~~~~~~~~~~~~~~~~~~~~~~~~~~~~~~i=1,2,\cdots,N+1,\\
		&\!\!\hat{\mathcal{X}}_l^{(q+1)}\!\!=\!\mathcal{P}_{\Omega^c_l}\!\bigg(\!\frac{\text{FCTN}\big(\!\{\!\mathcal{G}_i^{(q+1)}\!\}_{i=1}^{N+1}\!\big)\!\!+\!\!\rho\hat{\mathcal{X}}_l^{(q)}}{1\!+\!\rho}\!\bigg)\!\!+\!\!\mathcal{P}_{\Omega_l}(\hat{\mathcal{F}}_l).
	\end{aligned}
	\right.
	\label{solve_group}
\end{equation}

\begin{table*}[!t]
	\footnotesize
	\setlength{\tabcolsep}{6pt}
	\renewcommand\arraystretch{1.11}
	\caption{The average PSNR, SSIM, and SAM values on all testing MSIs of six utilized methods.}
	\begin{center}
		\begin{tabular}{c|ccc|ccc|ccc|ccc}
			\Xhline{0.8pt}
			
			MR          & \multicolumn{3}{c|}{98\%}          & \multicolumn{3}{c|}{95\%}          & \multicolumn{3}{c|}{90\%}          & \multicolumn{3}{c}{80\%}  \\
			
			\hline
			
			Method    & PSNR    & SSIM    & SAM    & PSNR    & SSIM    & SAM    & PSNR    & SSIM    & SAM    & PSNR    & SSIM    & SAM    \\
			
			\hline
			
			KBR-TC    & 22.6409    & 0.6428    & 20.4000    & 33.5480    & 0.9110    & 6.8097    & 41.7296    & 0.9774    & 3.4463    & 48.5367    & 0.9941    & 1.9720    \\
			NL-SNN    & 16.5703    & 0.6420    & 17.9006    & 29.6485    & 0.8645    & 6.1118    & 41.3033    & 0.9774    & 2.5377    & 47.3655    & 0.9914    & 1.7127    \\
			NL-TNN    & 19.8220    & 0.6568    & 26.3135    & 31.6988    & 0.9219    & 8.6298    & 36.7736    & 0.9704    & 3.9475    & 44.1860    & 0.9930    & 1.7618    \\
			NL-TT     & 26.3712    & 0.7862    & 15.2674    & 31.4861    & 0.9051    & 6.6808    & 34.7360 	  & 0.9472    & 3.9148    & 38.4261    & 0.9752    & 2.5169    \\
			FCTN-TC   & 25.9944    & 0.6625    & 19.1886    & 33.0224    & 0.8631    & 9.8324    & 37.7269    & 0.9373    & 6.6739    & 43.6822    & 0.9760    & 4.2126    \\
			NL-FCTN   & \textbf{28.4943}    & \textbf{0.8042}    & \textbf{11.9528}    & \textbf{37.5379}    & \textbf{0.9596}    & \textbf{3.9858}    & \textbf{44.2862}    & \textbf{0.9901}    & \textbf{2.0703}    & \textbf{49.7174}    & \textbf{0.9964}    & \textbf{1.3360}    \\ 
			
			\Xhline{0.8pt}
		\end{tabular}
	\end{center}
	\label{mHSItab}
	\vspace{-0.5em}
\end{table*}

\begin{figure*}[!t]
	\footnotesize
	\setlength{\tabcolsep}{1pt}
	\begin{center}
		\begin{tabular}{cccccccc}
			\includegraphics[width=0.12\textwidth]{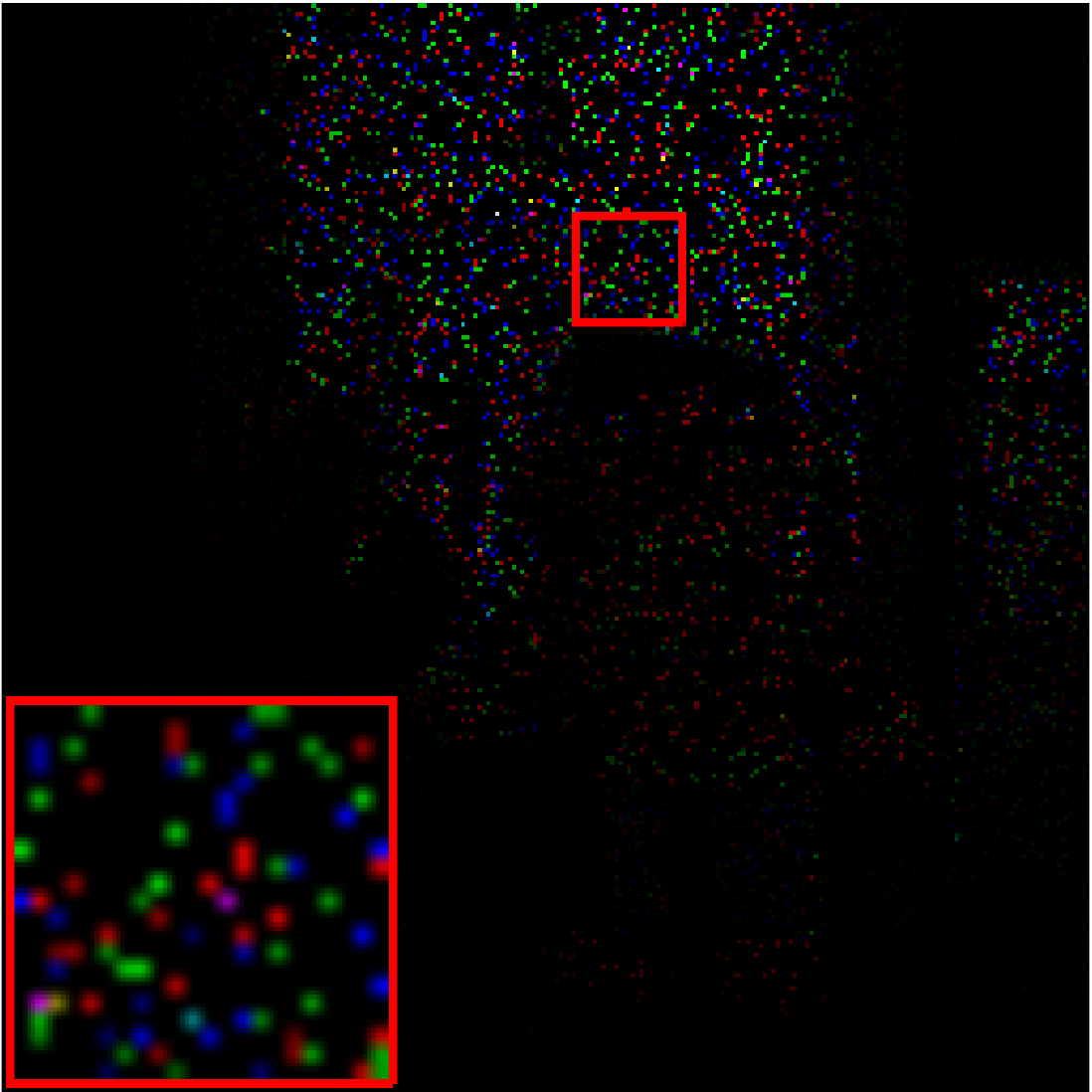}&
			\includegraphics[width=0.12\textwidth]{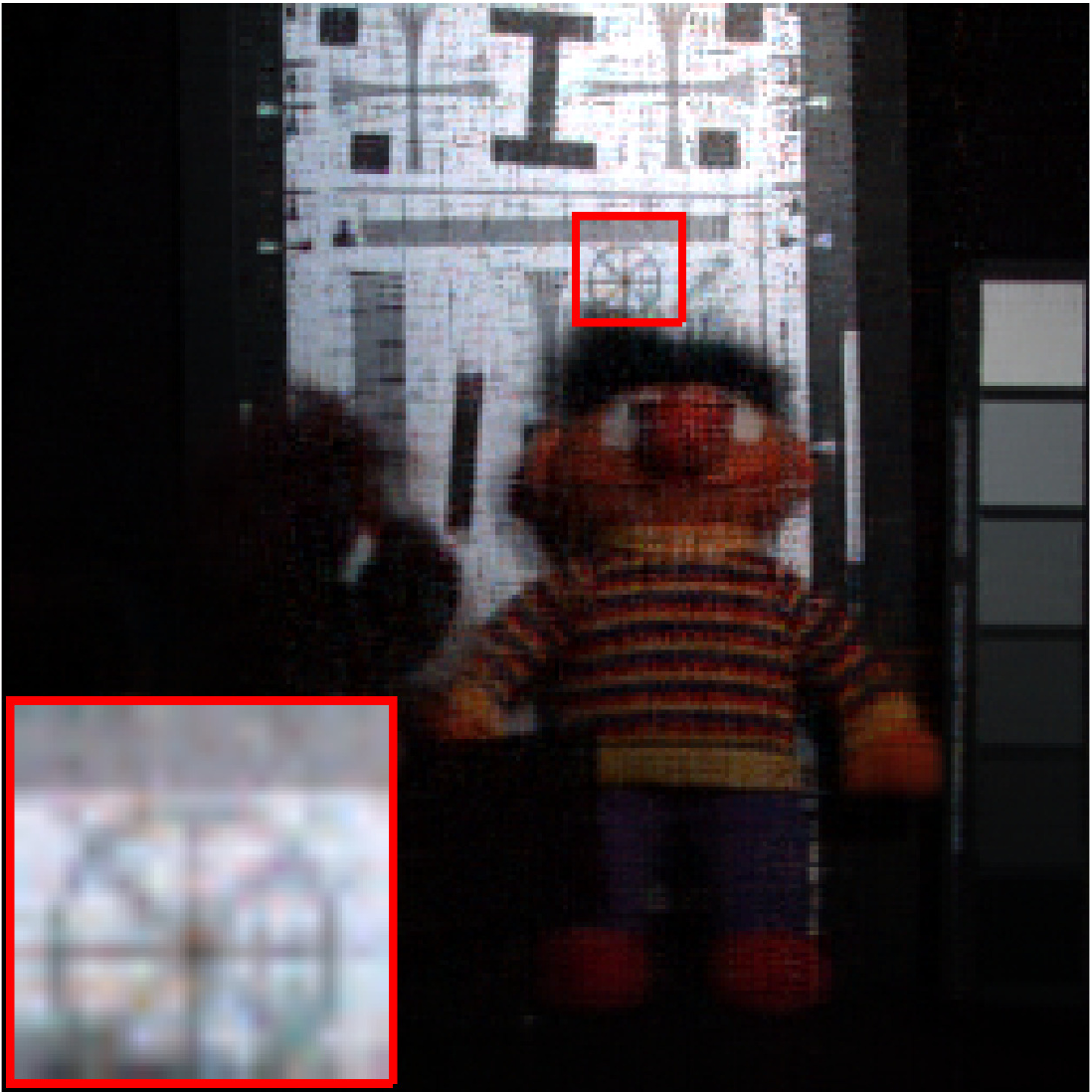}&
			\includegraphics[width=0.12\textwidth]{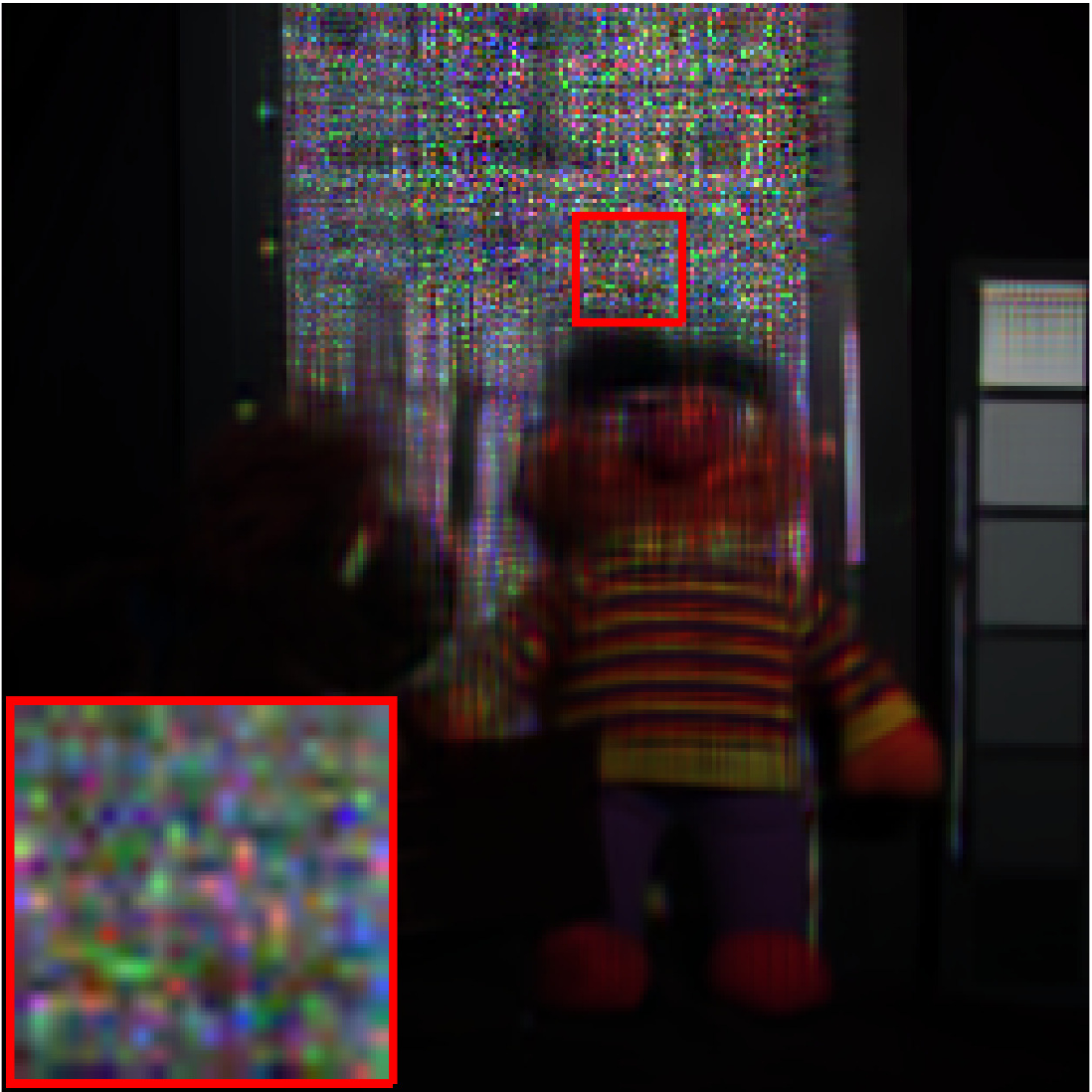}&
			\includegraphics[width=0.12\textwidth]{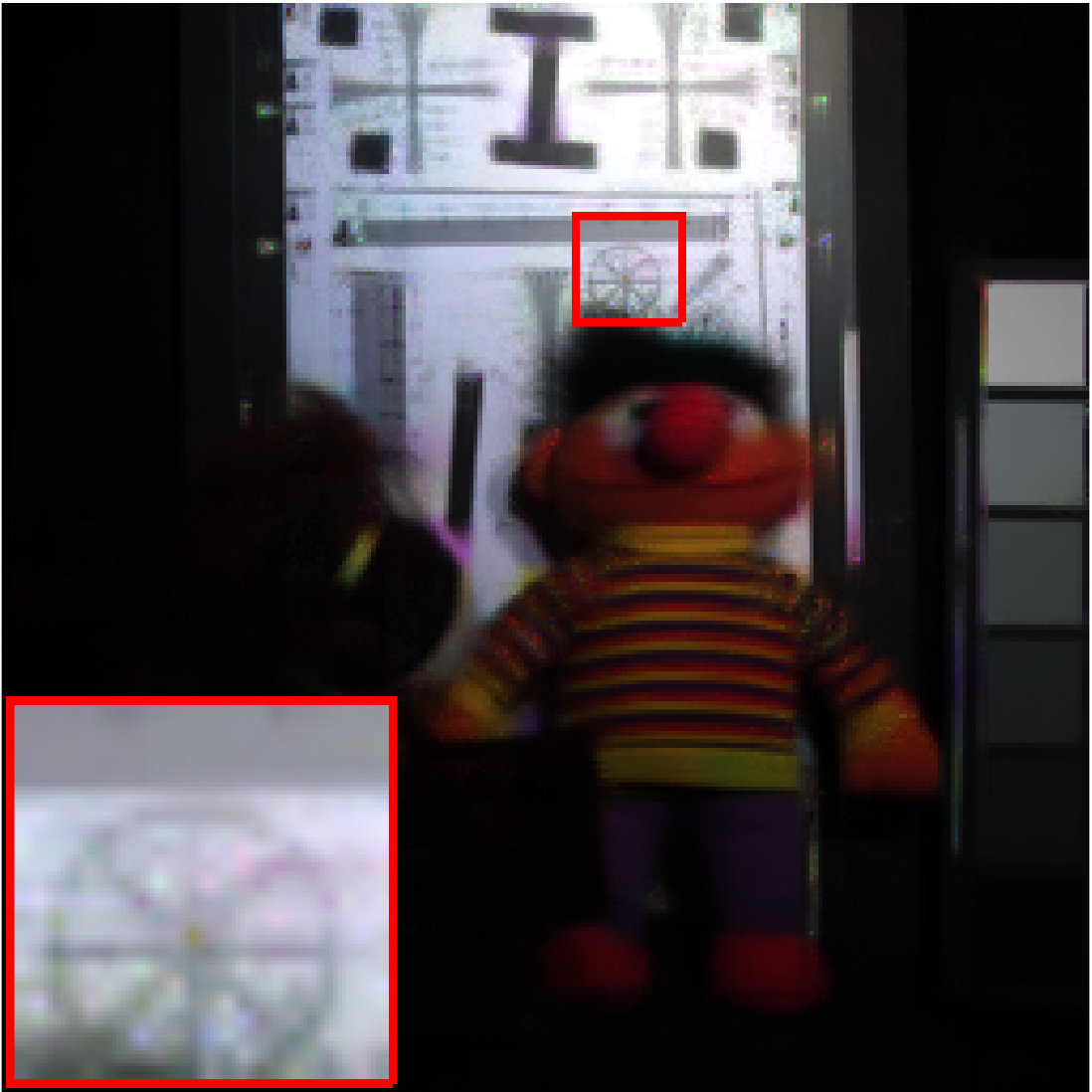}&
			\includegraphics[width=0.12\textwidth]{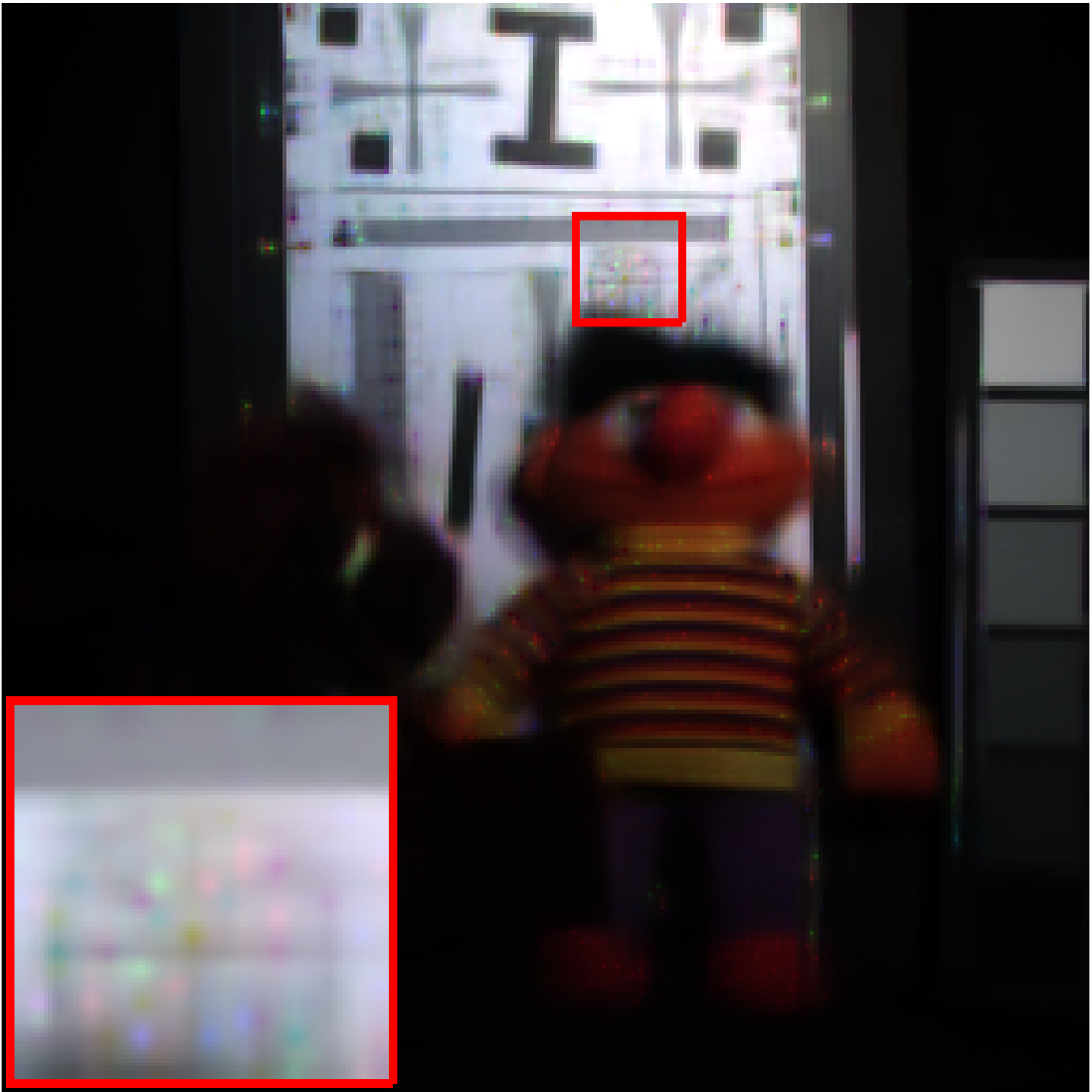}&
			\includegraphics[width=0.12\textwidth]{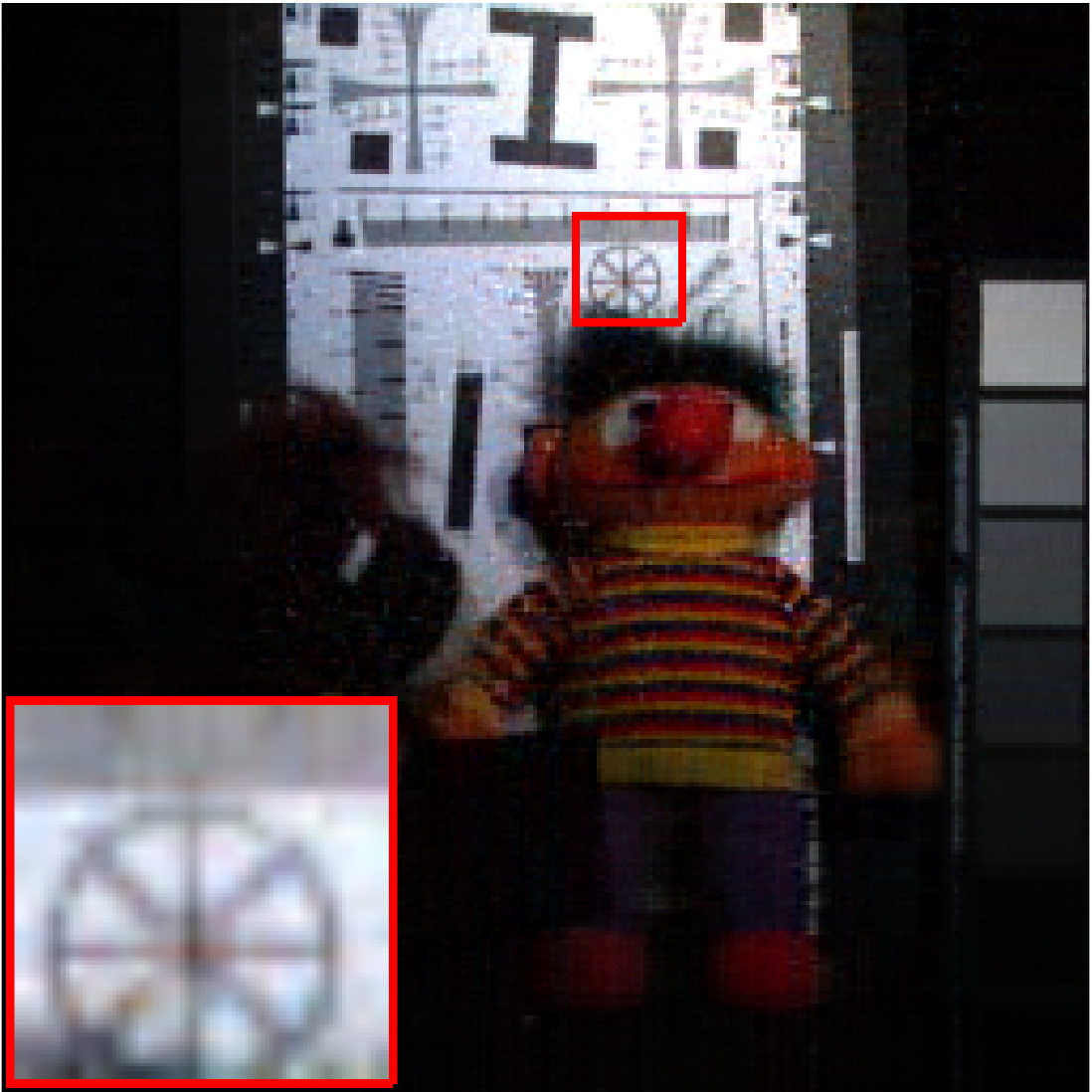}&
			\includegraphics[width=0.12\textwidth]{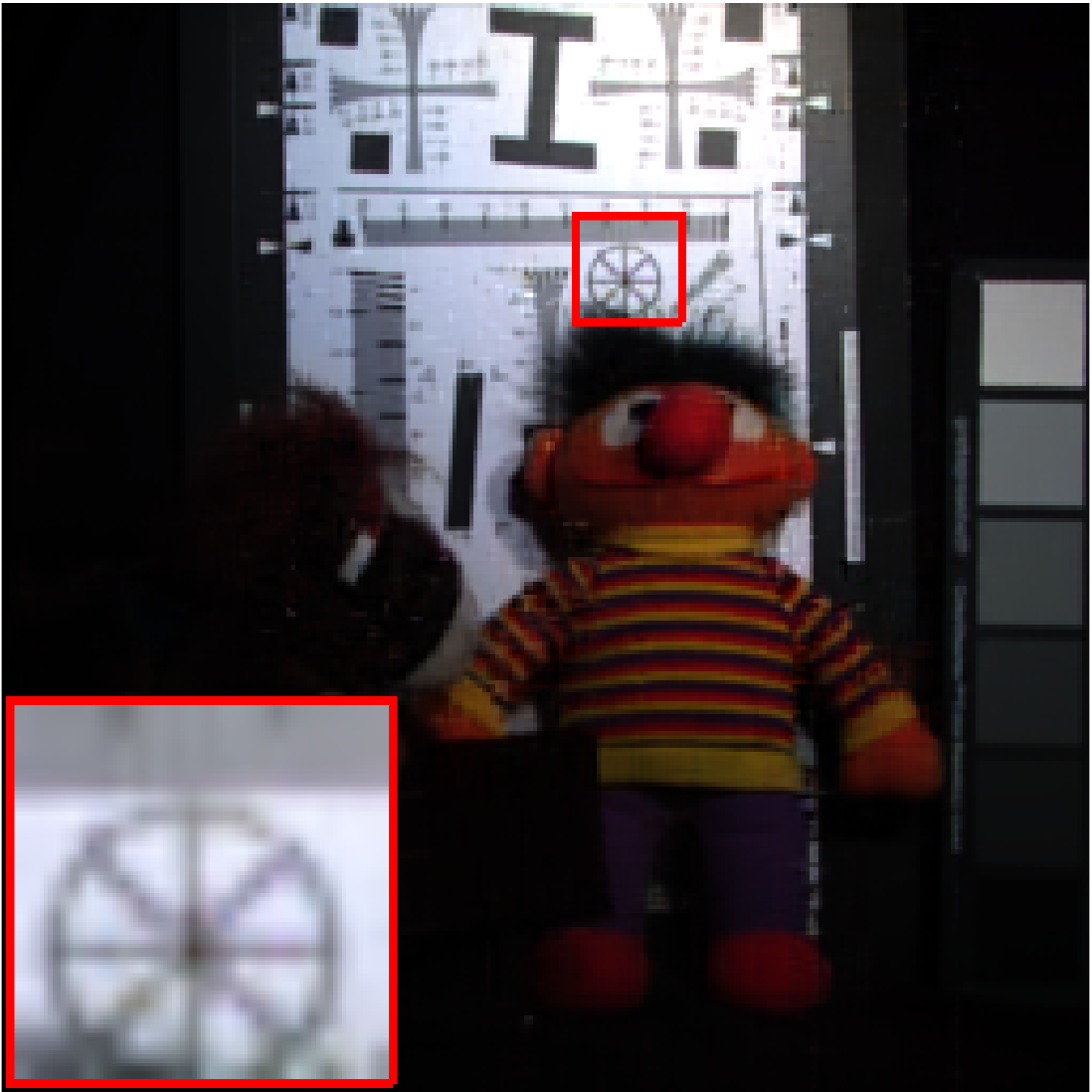}&
			\includegraphics[width=0.12\textwidth]{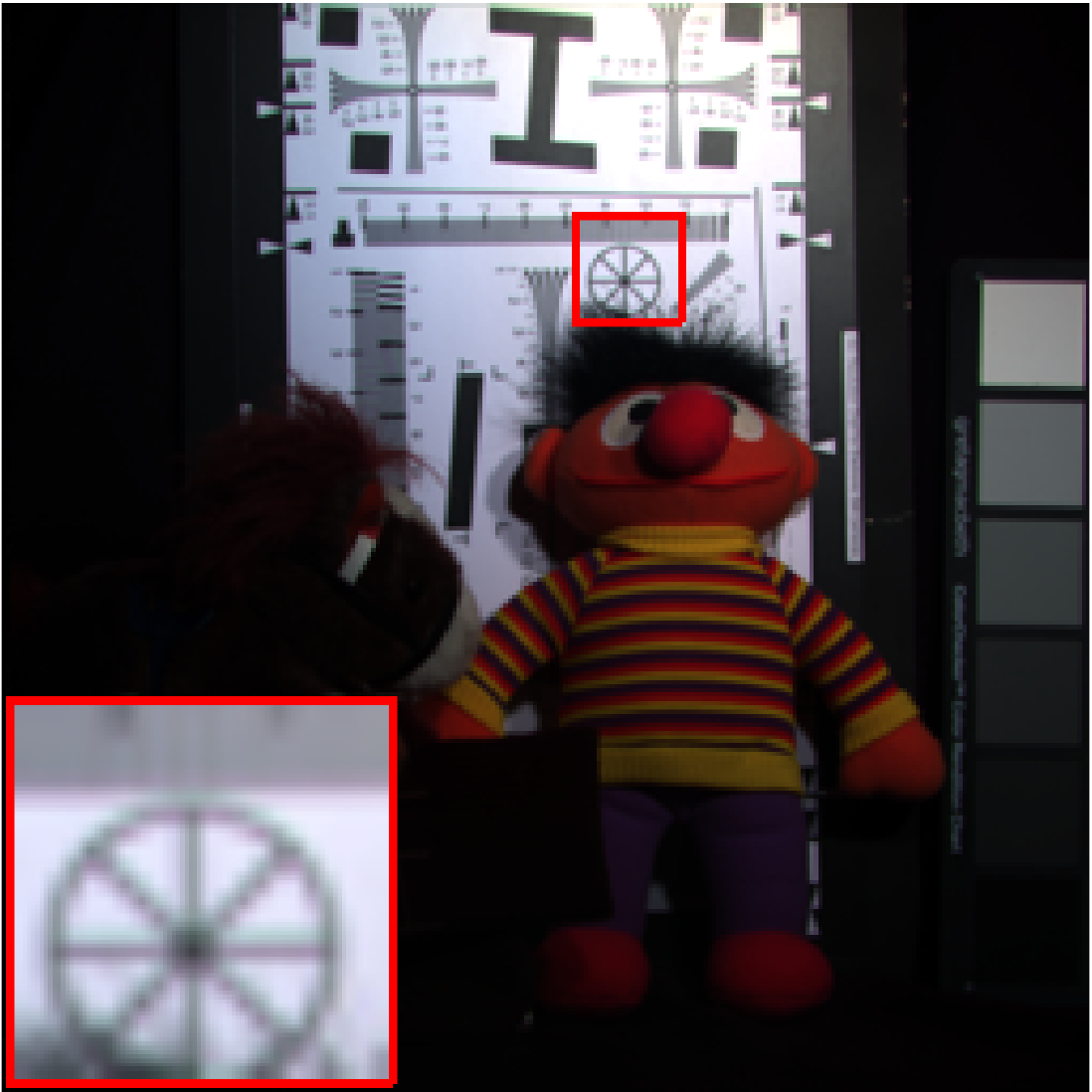}\\
			\includegraphics[width=0.12\textwidth]{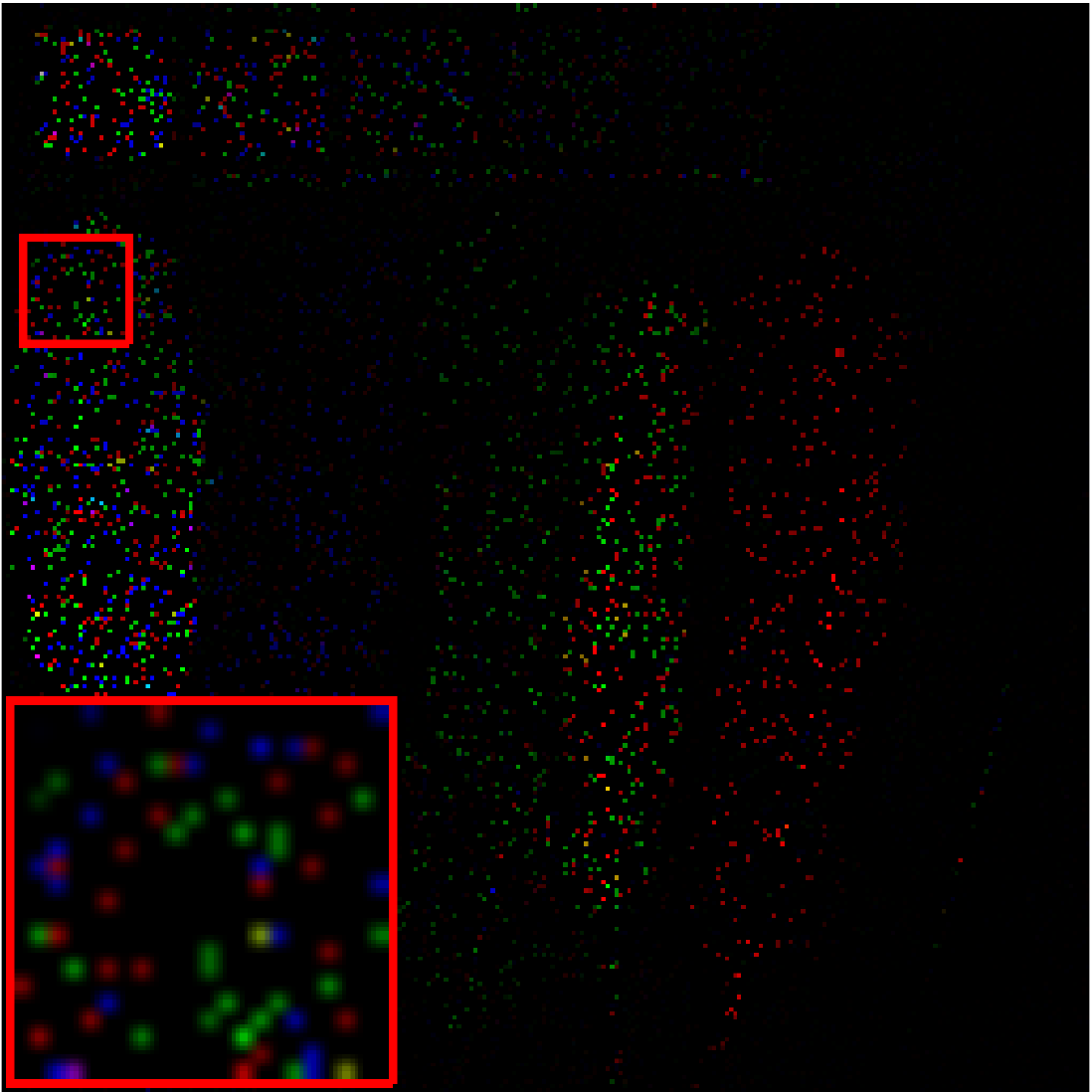}&
			\includegraphics[width=0.12\textwidth]{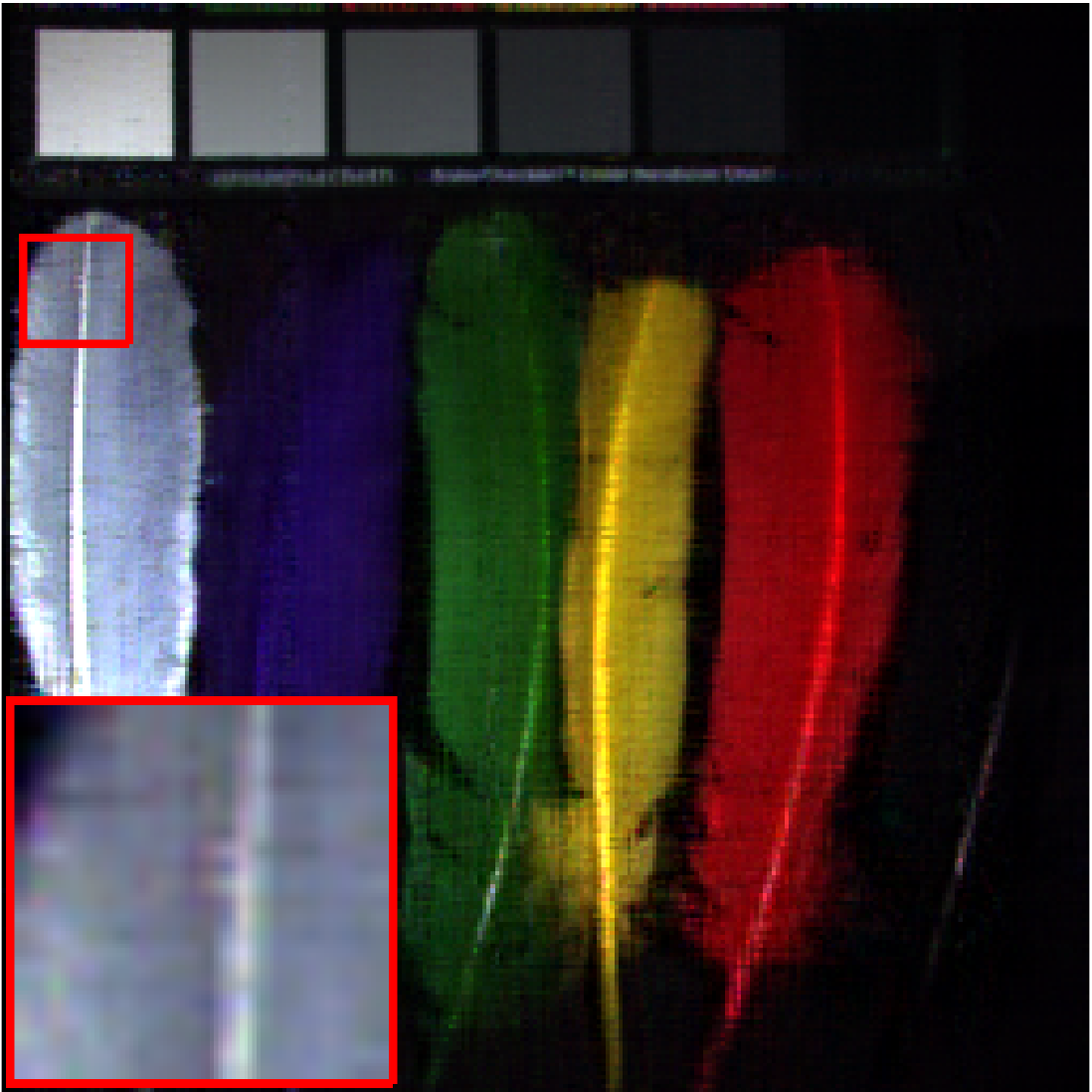}&
			\includegraphics[width=0.12\textwidth]{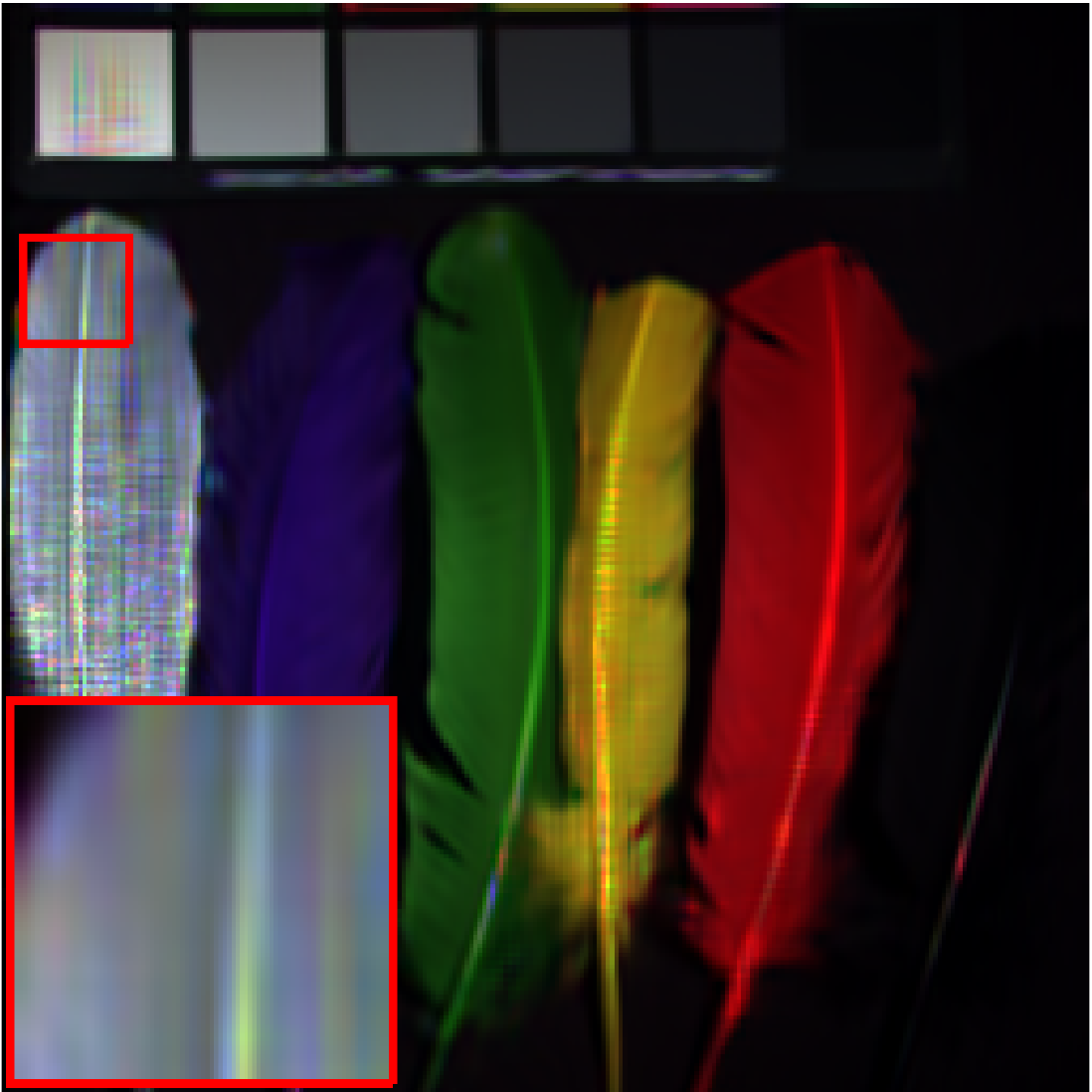}&
			\includegraphics[width=0.12\textwidth]{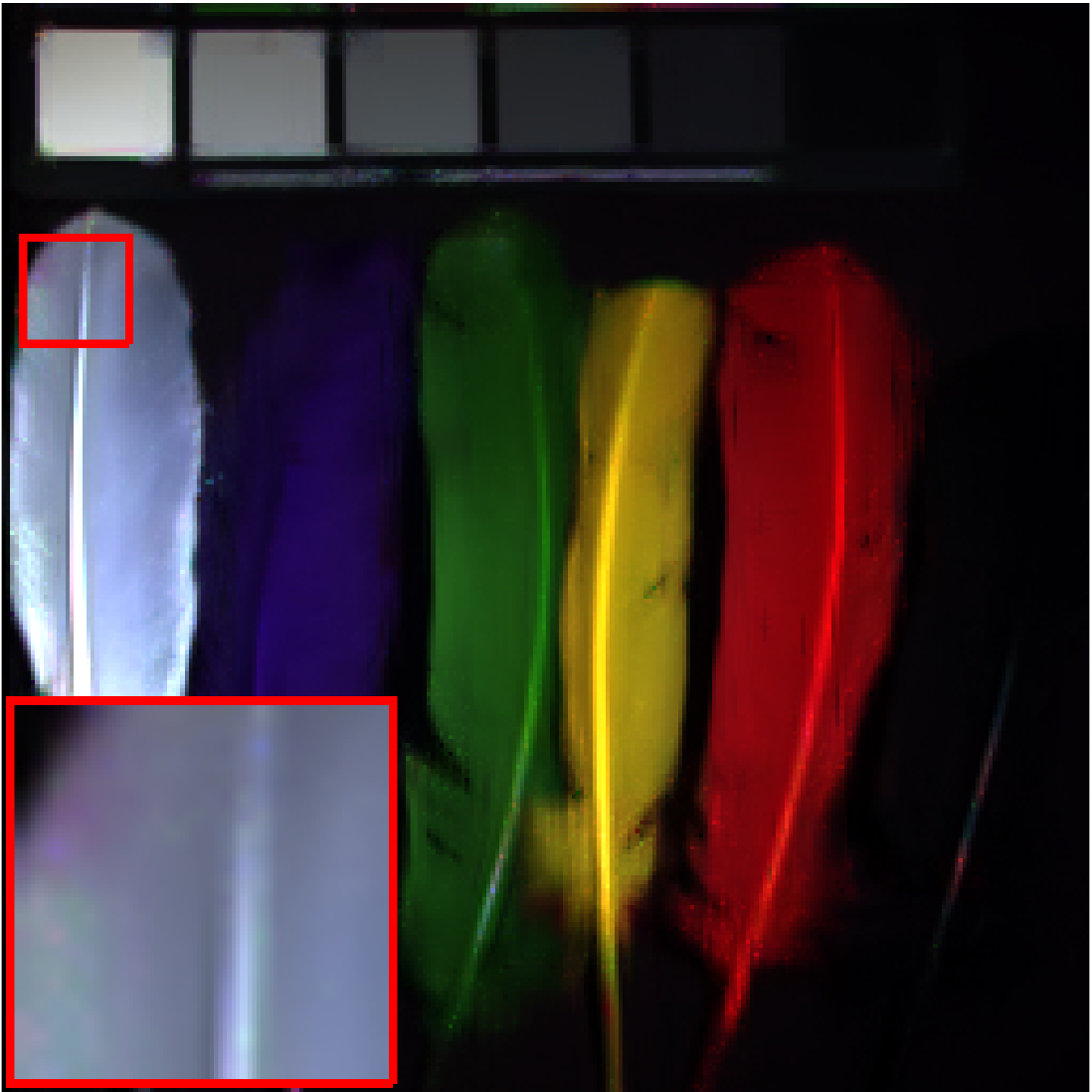}&
			\includegraphics[width=0.12\textwidth]{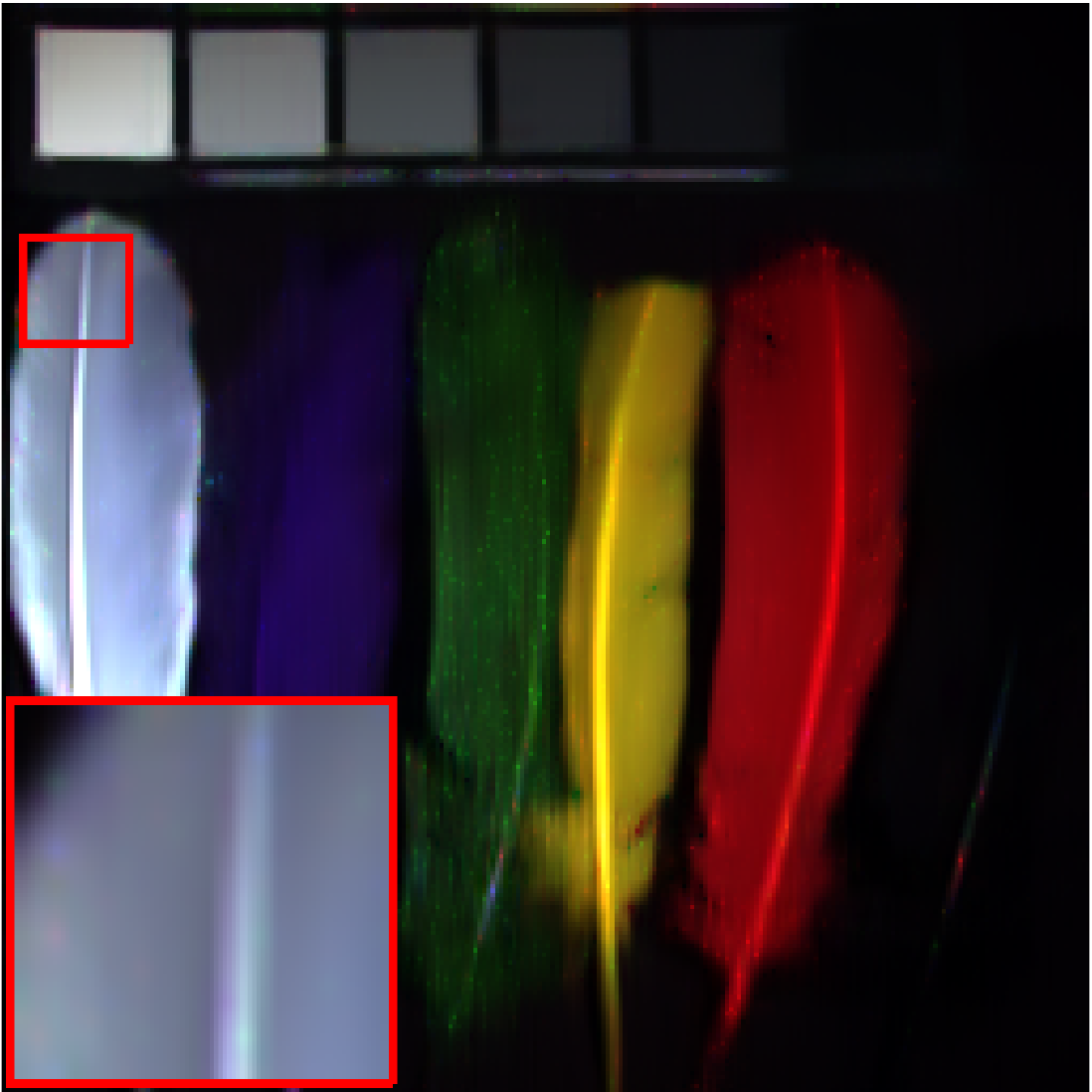}&
			\includegraphics[width=0.12\textwidth]{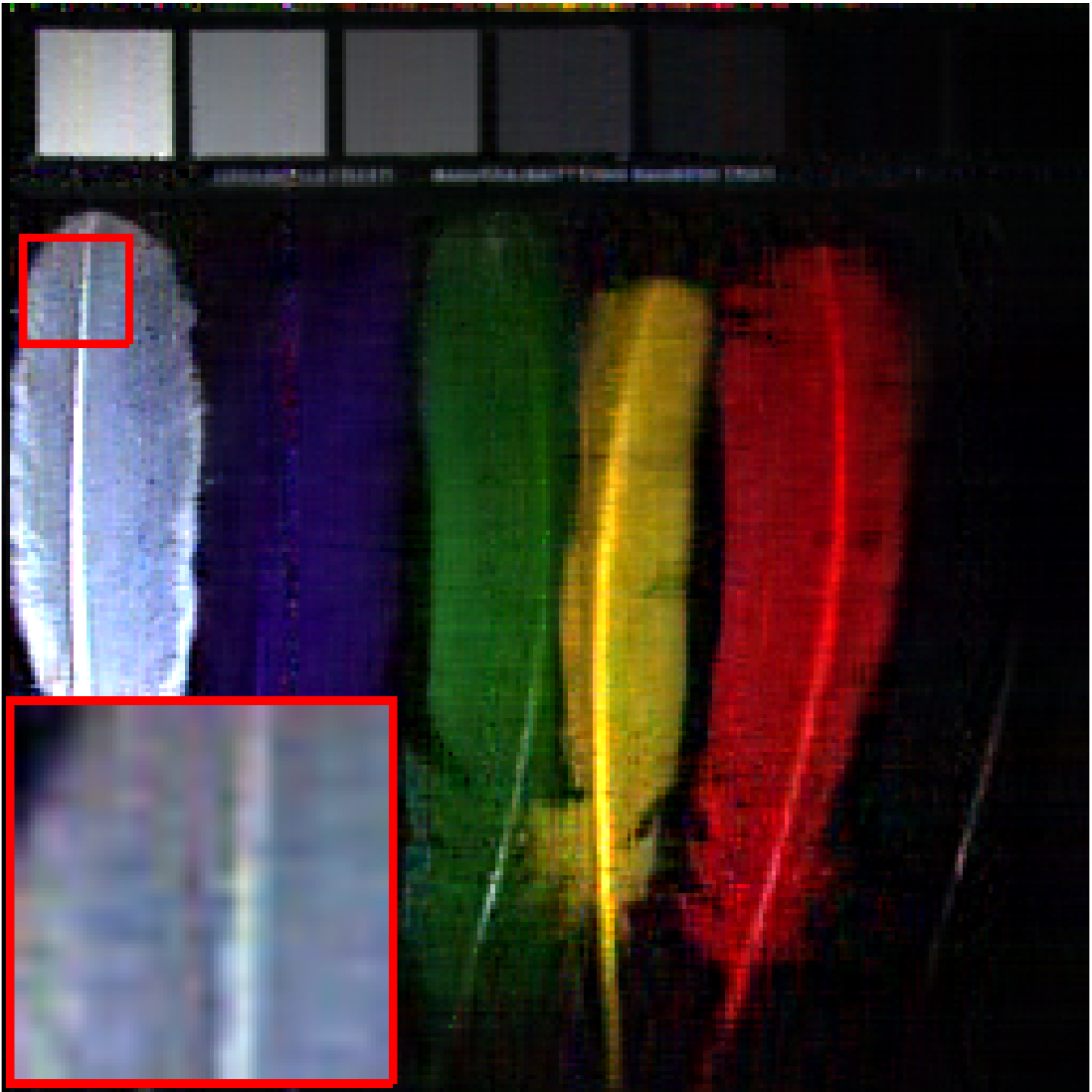}&
			\includegraphics[width=0.12\textwidth]{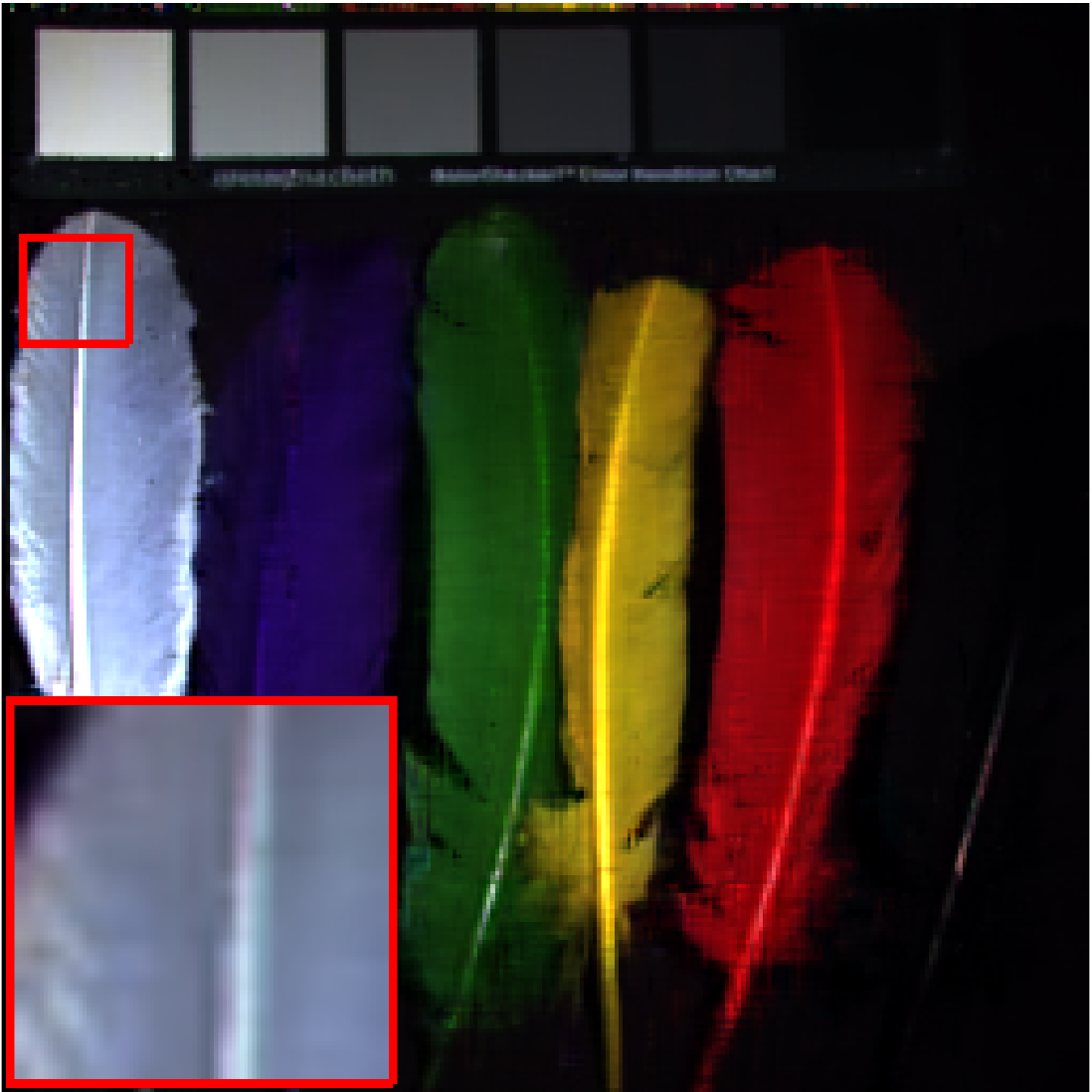}&
			\includegraphics[width=0.12\textwidth]{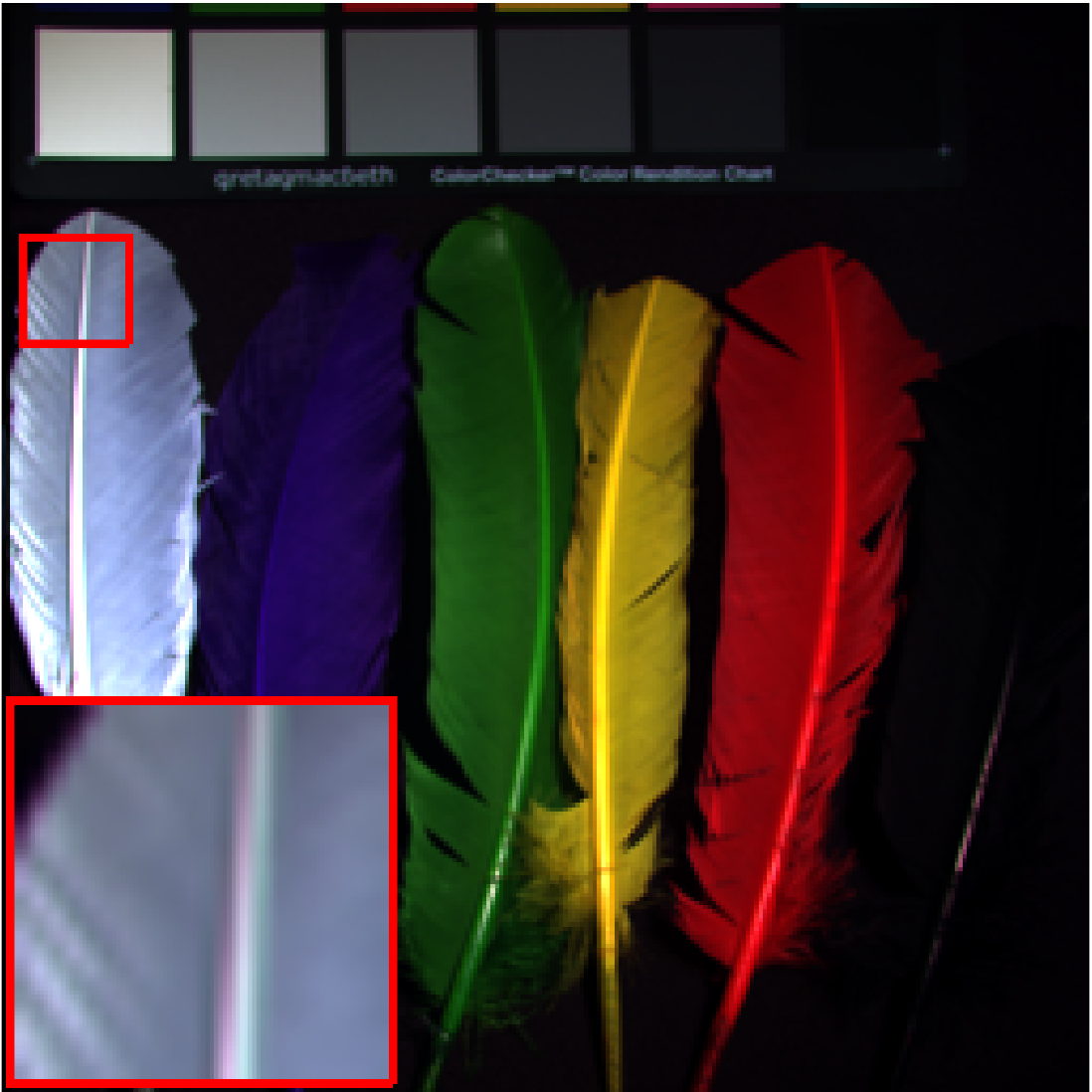}\\
			\includegraphics[width=0.12\textwidth]{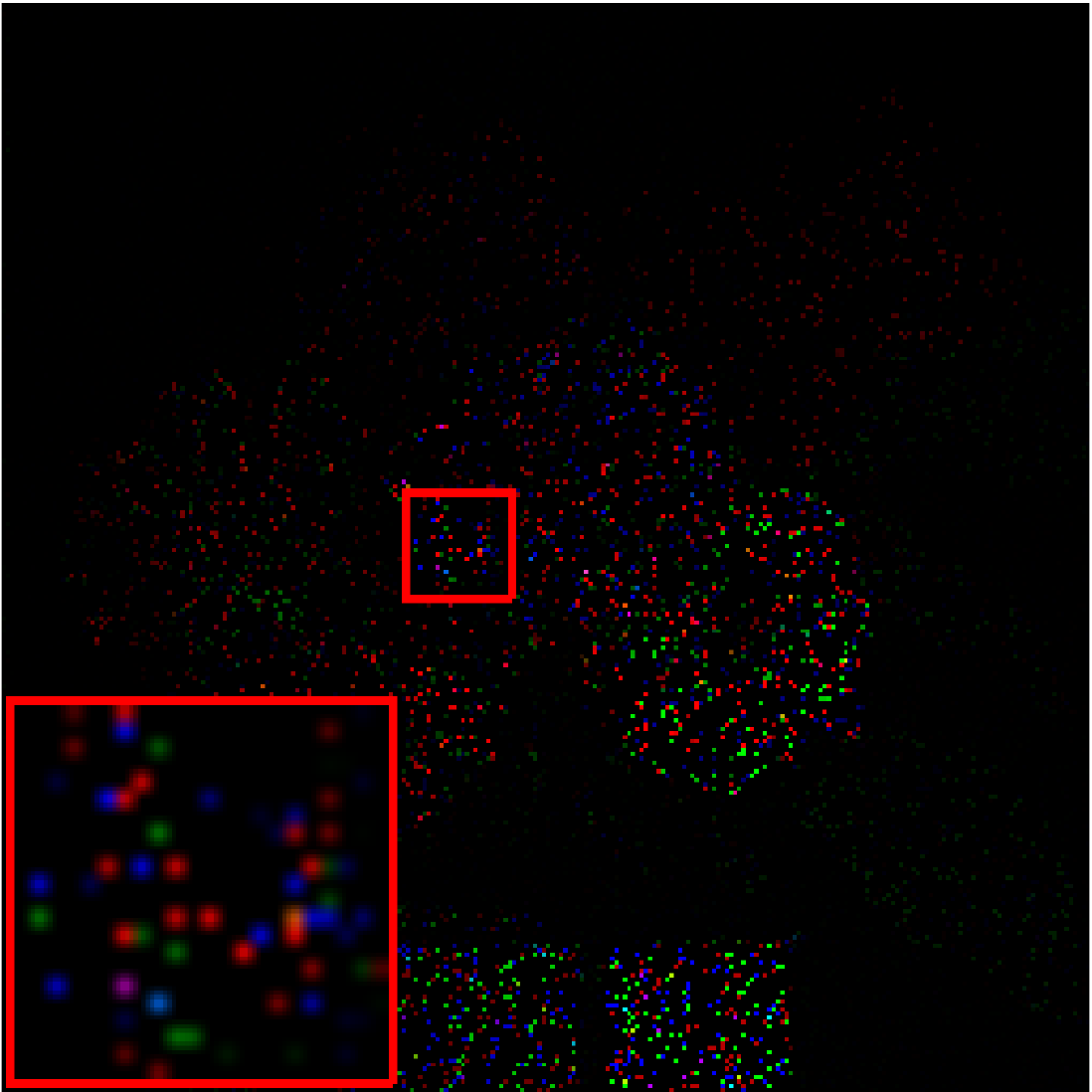}&
			\includegraphics[width=0.12\textwidth]{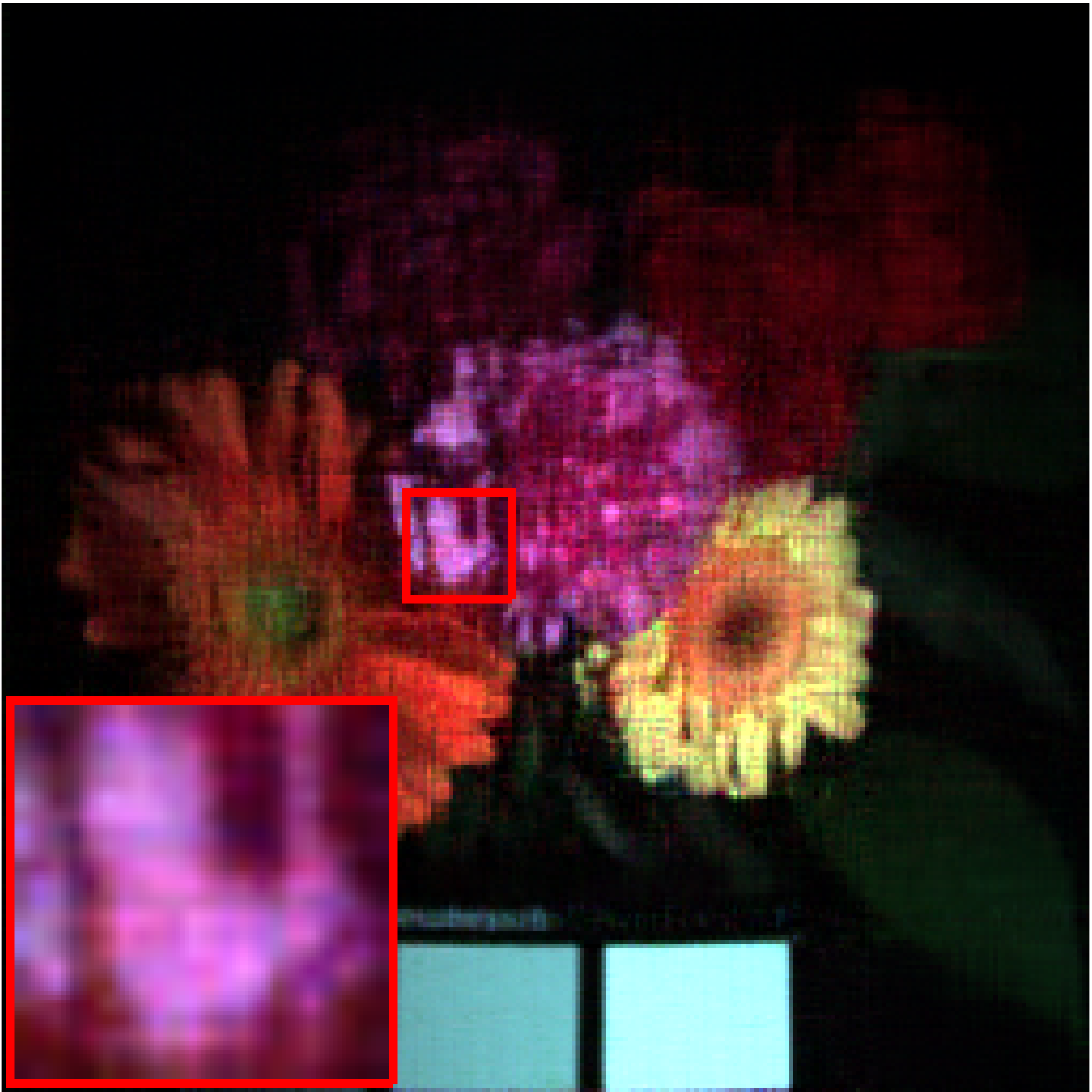}&
			\includegraphics[width=0.12\textwidth]{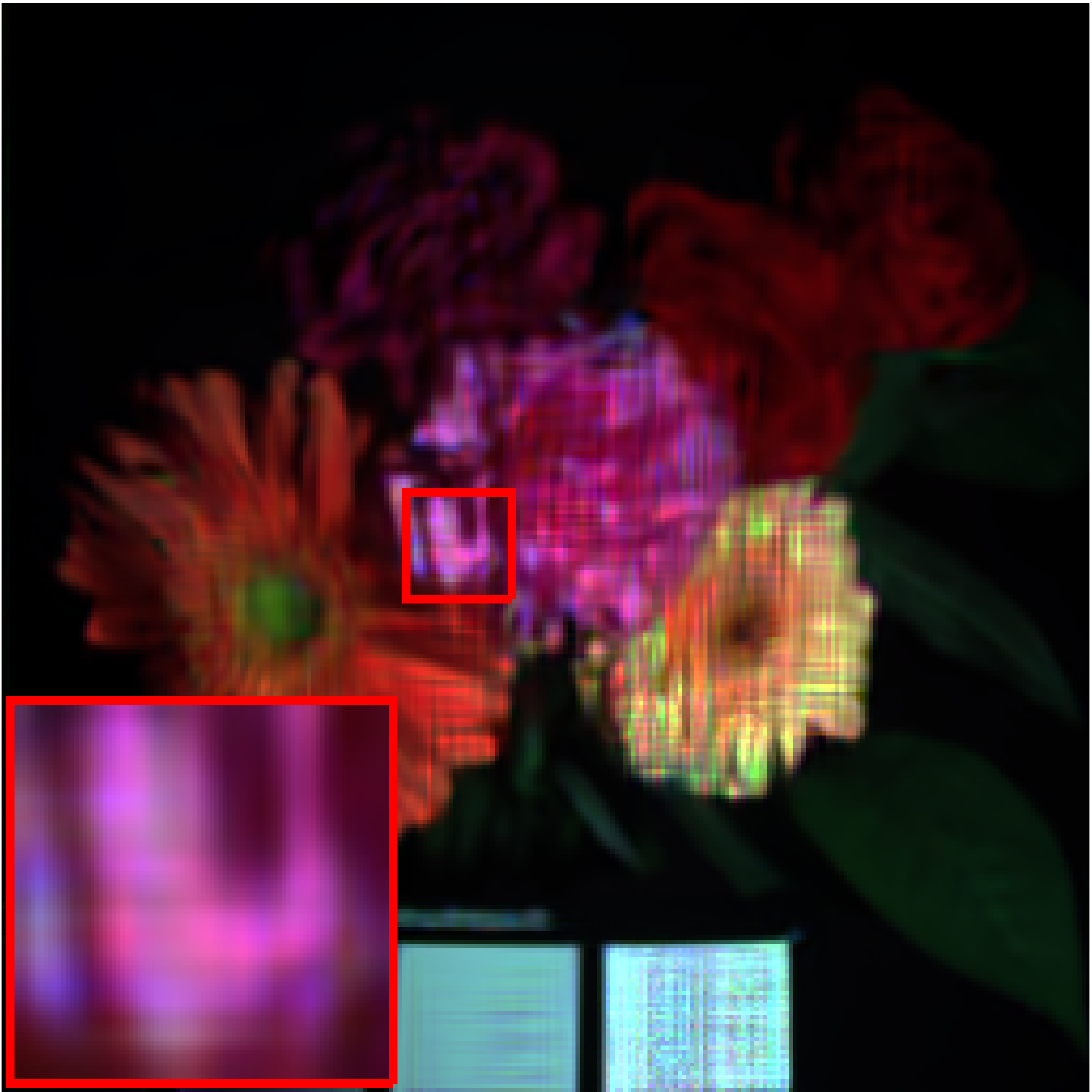}&
			\includegraphics[width=0.12\textwidth]{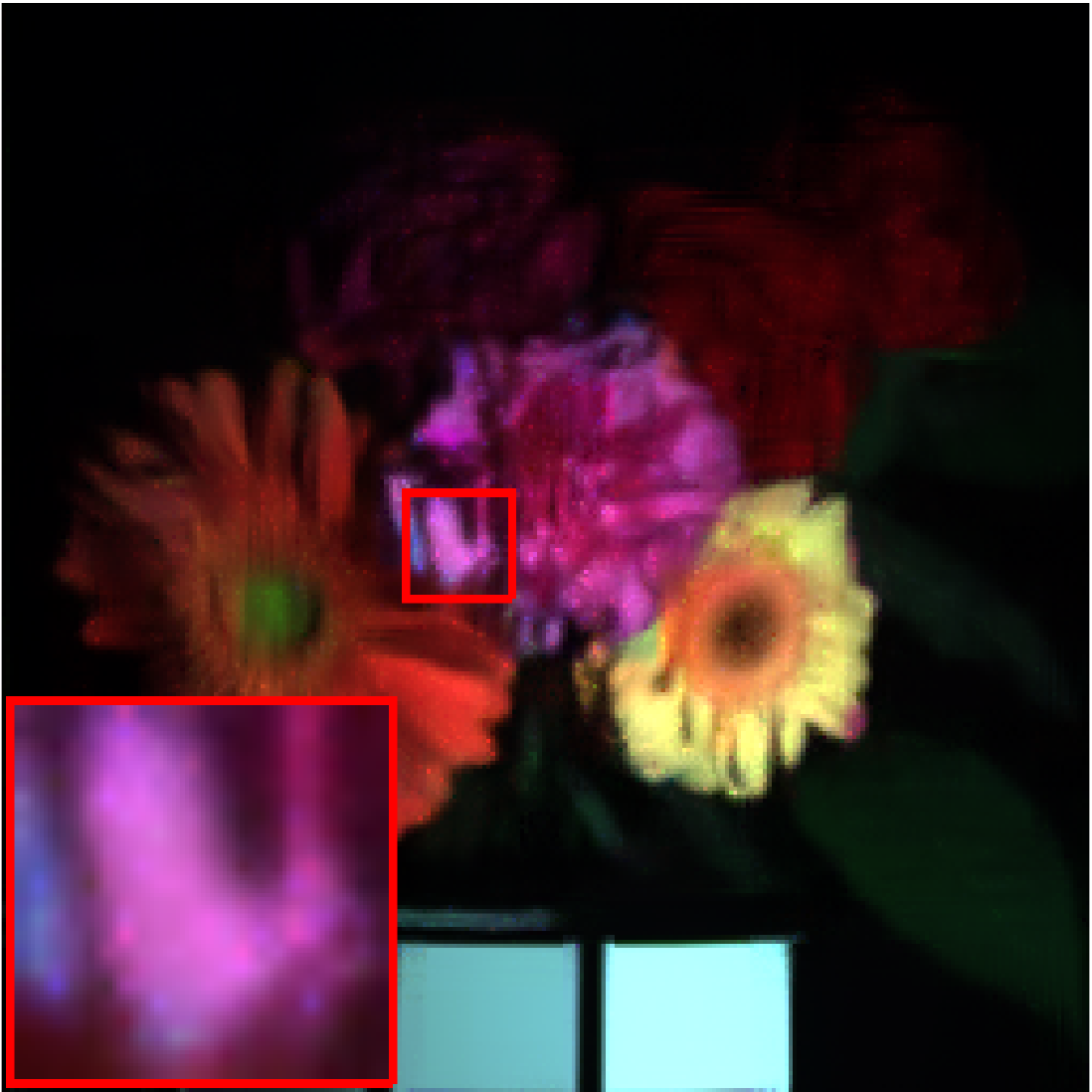}&
			\includegraphics[width=0.12\textwidth]{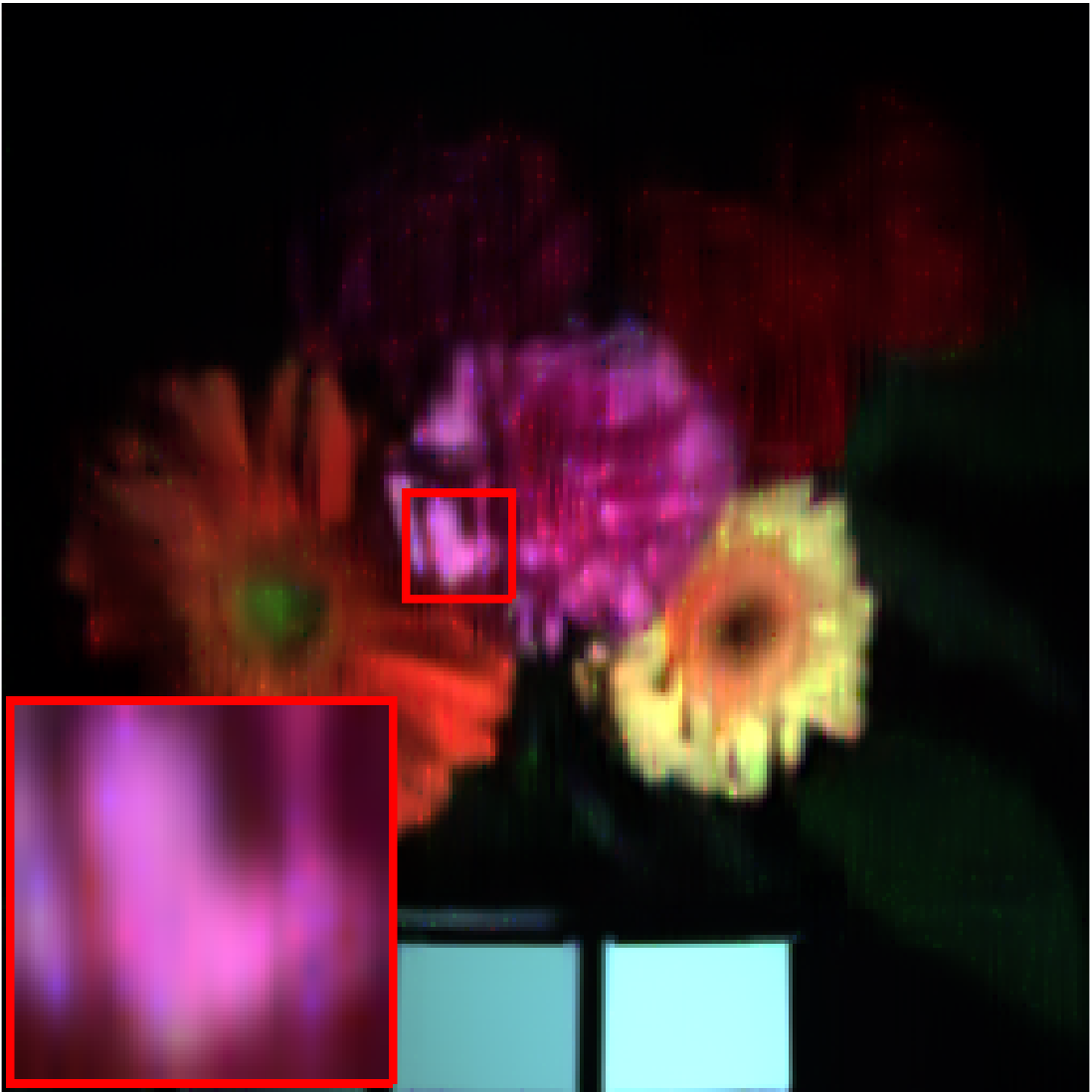}&
			\includegraphics[width=0.12\textwidth]{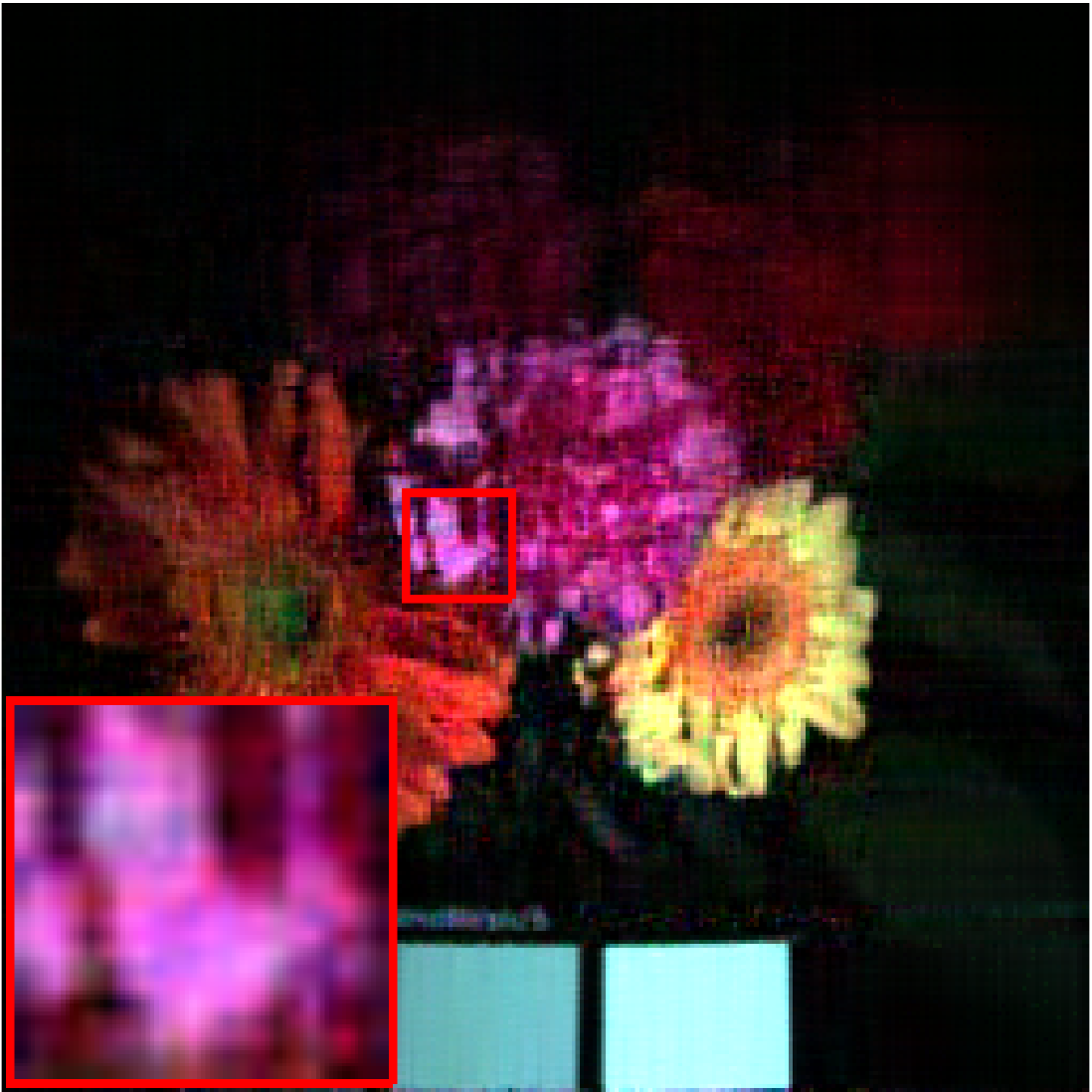}&
			\includegraphics[width=0.12\textwidth]{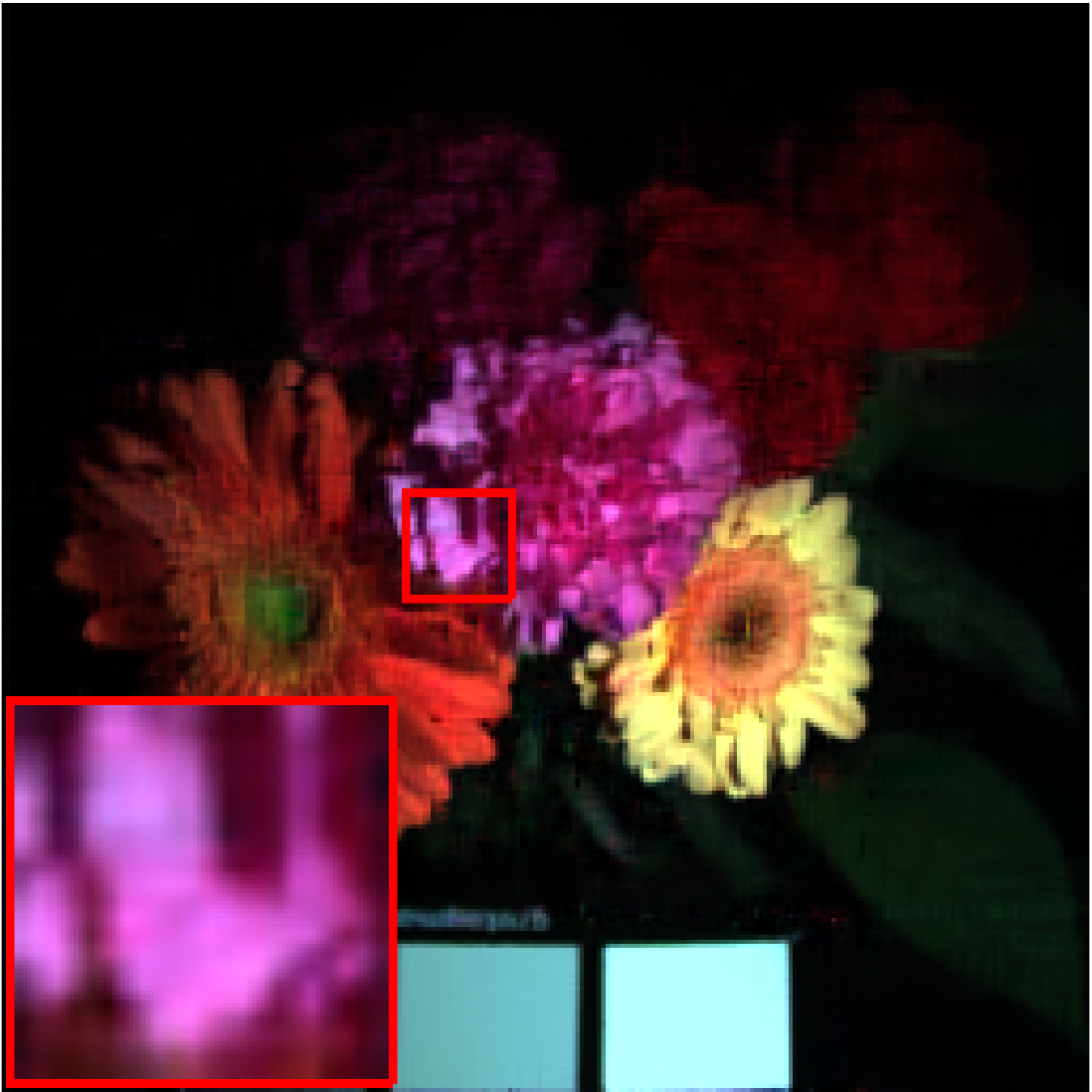}&
			\includegraphics[width=0.12\textwidth]{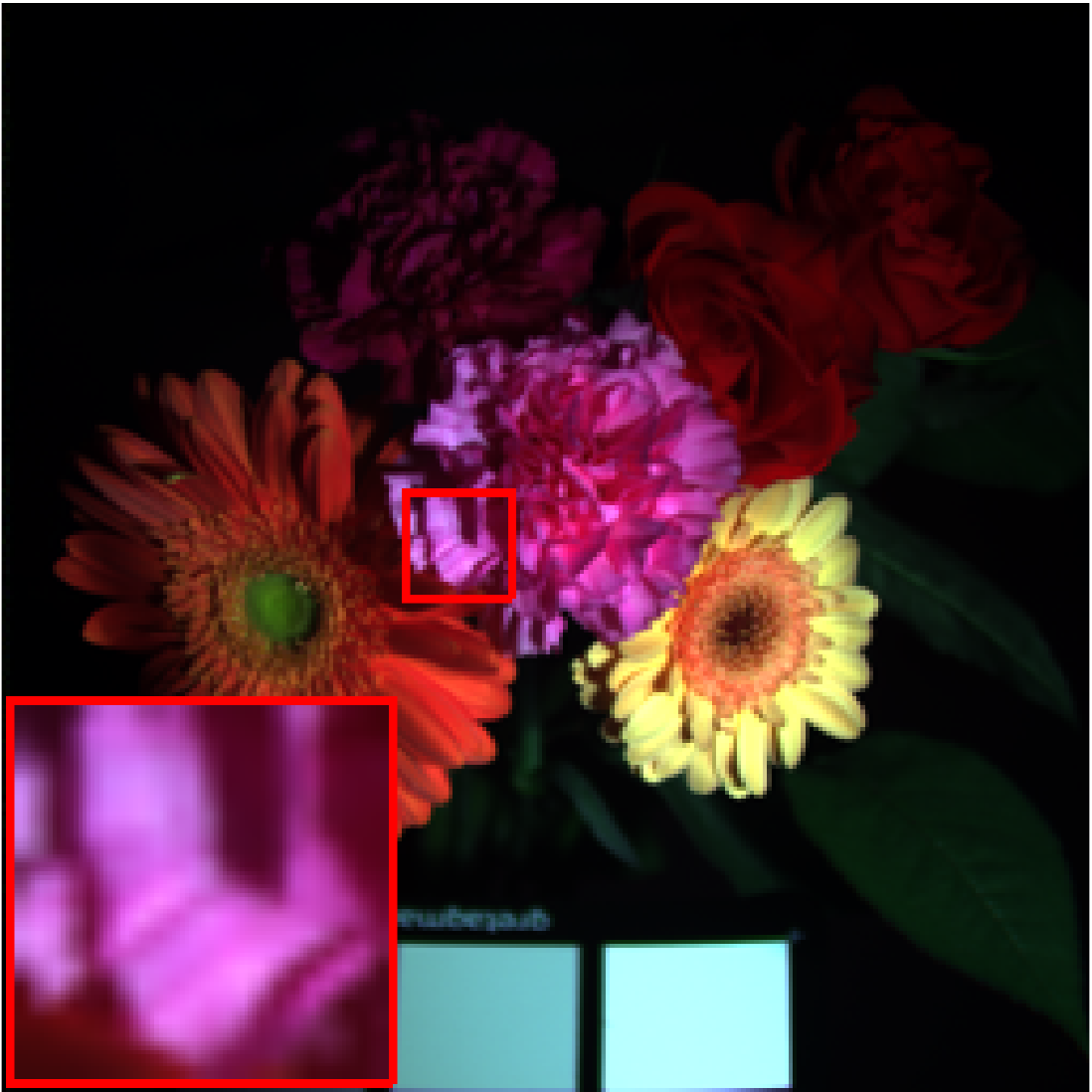}\\[0.05cm]
			Degraded image  & KBR-TC  & NL-SNN  & NL-TNN  & NL-TT  & FCTN-TC  & NL-FCTN  & Original image
		\end{tabular}
	\end{center}
	\caption{Restoration results of three selected MSIs with MR=95\%. The pseudo color images are composed of bands 29, 14, and 5. From top to bottom: the results of \emph{chart and stuffed toy}, \emph{feathers}, and \emph{flowers}.}\label{mHSIfig}
	\vspace{-1.2em}
\end{figure*}

In step B.III, we aggregate the inpainted NSS groups $\{\hat{\mathcal{X}}_l\}^L_{l=1}$ to their original positions and obtain the whole inpainted RSI $\mathcal{X}$. 

The above NL-FCTN decomposition-based method for RSI inpainting is summarized in Algorithm \ref{algorithms_NLFCTN}. Its theoretical convergence can be similarly proved following \cite{zheng2021fully}.

\section{Numerical Experiments}\label{exper}

In this section, we test the performance of NL-FCTN decomposition-based method for RSI inpainting on the MSIs (third-order tensors) and the time-series sentinel-2 images (fourth-order tensors). The compared methods are global tensor decomposition-based tensor completion (TC) methods including KBR-TC\cite{xie2017kronecker} and FCTN-TC\cite{zheng2021fully}, and nonlocal patch-based tensor decomposition for TC including NL-SNN\cite{xie2018tensor}, NL-TNN\cite{song2018nonlocal}, and NL-TT\cite{ding2021tensor}. All hyper-parameters of each compared method are set for the optimal performance according to the authors' paper and code. Three quantitative metrics contain the peak signal to noise rate (PSNR), the structural similarity (SSIM), and the spectral angle mapping (SAM).

\emph{1) MSI Inpainting:} We test the CAVE Database\footnotemark[1]\footnotetext[1]{https://www.cs.columbia.edu/CAVE/databases/multispectral/.} with 32 MSIs. Each test MSI is resized to 256 $\times$ 256 $\times$ 31 and rescaled to [0,1]. The missing rate (MR) is defined as the rate of the unknown elements in the total elements. We test four MRs: 98\%, 95\%, 90\%, and 80\%. Table \ref{mHSItab} reports the average values of PSNR, SSIM, and SAM on all 32 MSIs by six compared methods under each MR. We observe that the proposed method performs better than the compared methods for all quantitative metrics. Particularly, our method is around 4 dB higher than the second best performed method with MR=95\%. Compared with other methods, we also observe that KBR-TC, NL-SNN, and NL-TNN perform well with low MR and perform poorly with high MR; NL-TT performs poorly with low MR and performs well with high MR; FCTN-TC shows moderate but stable performance. Naturally, NL-FCTN introduces FCTN-TC to the whole MSI and its NSS groups, which achieves the best performance.

Fig. \ref{mHSIfig} shows the pseudo color images of restoration results with MR=95\% of three MSIs. As observed, the proposed method achieves superior inpainting results visually. In particular, the spatial textures of MSIs can be better preserved, such as the lines in \emph{feathers} and the petals in \emph{flowers}.

\begin{figure*}[!t]
	\setlength{\abovecaptionskip}{0.cm}
	\footnotesize
	\setlength{\tabcolsep}{1pt}
	\begin{center}
		\begin{tabular}{cccccccc}
			\includegraphics[width=0.12\textwidth]{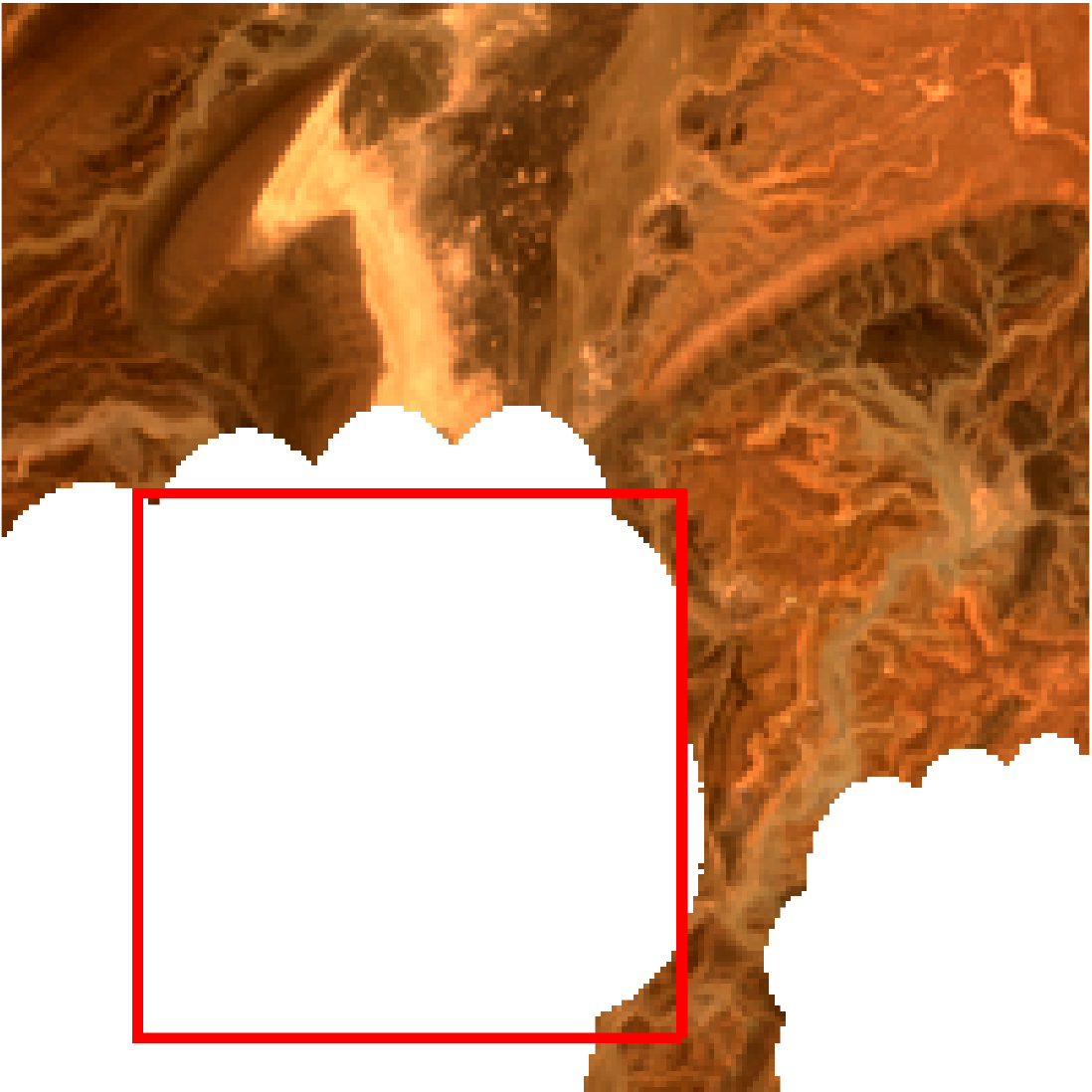}&
			\includegraphics[width=0.12\textwidth]{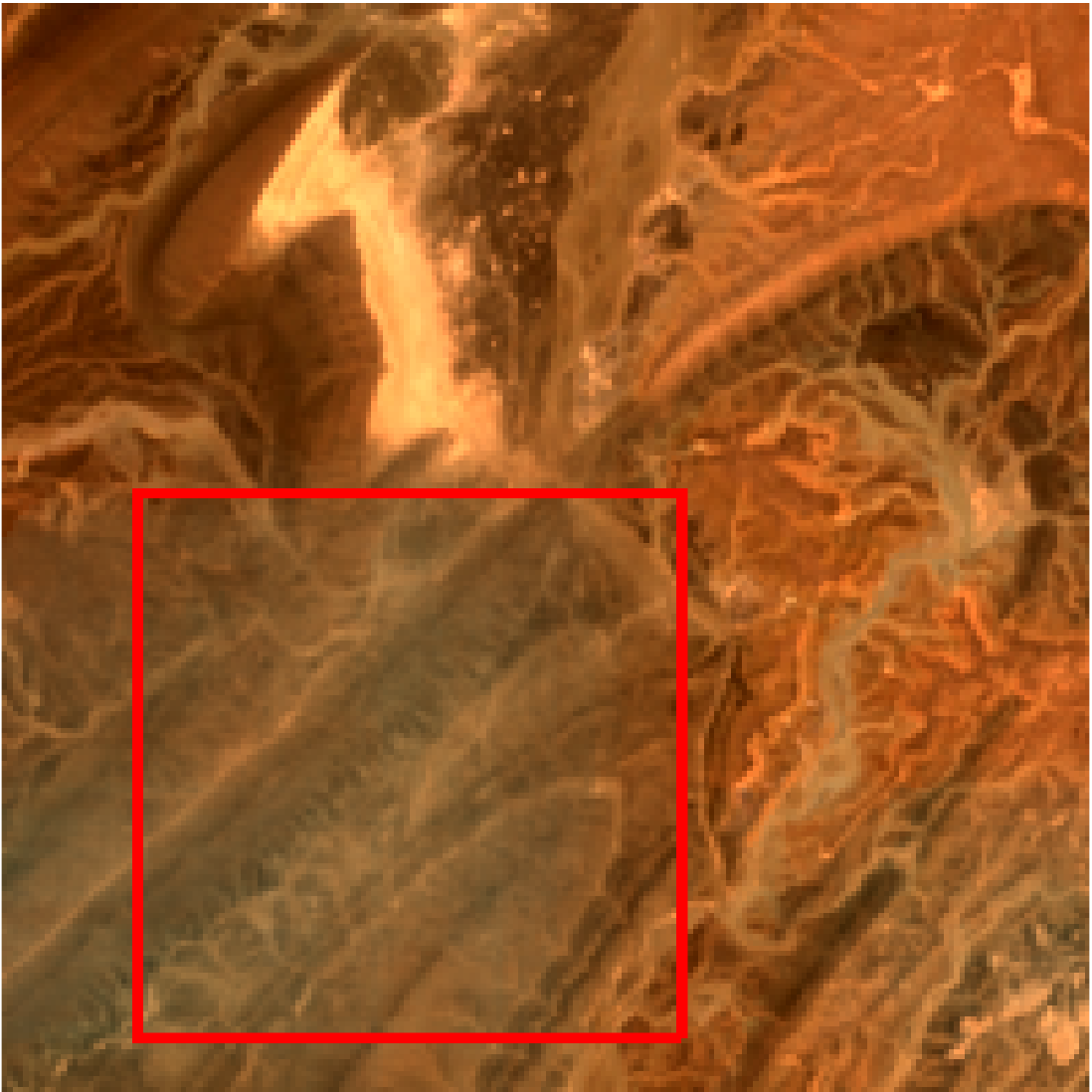}&
			\includegraphics[width=0.12\textwidth]{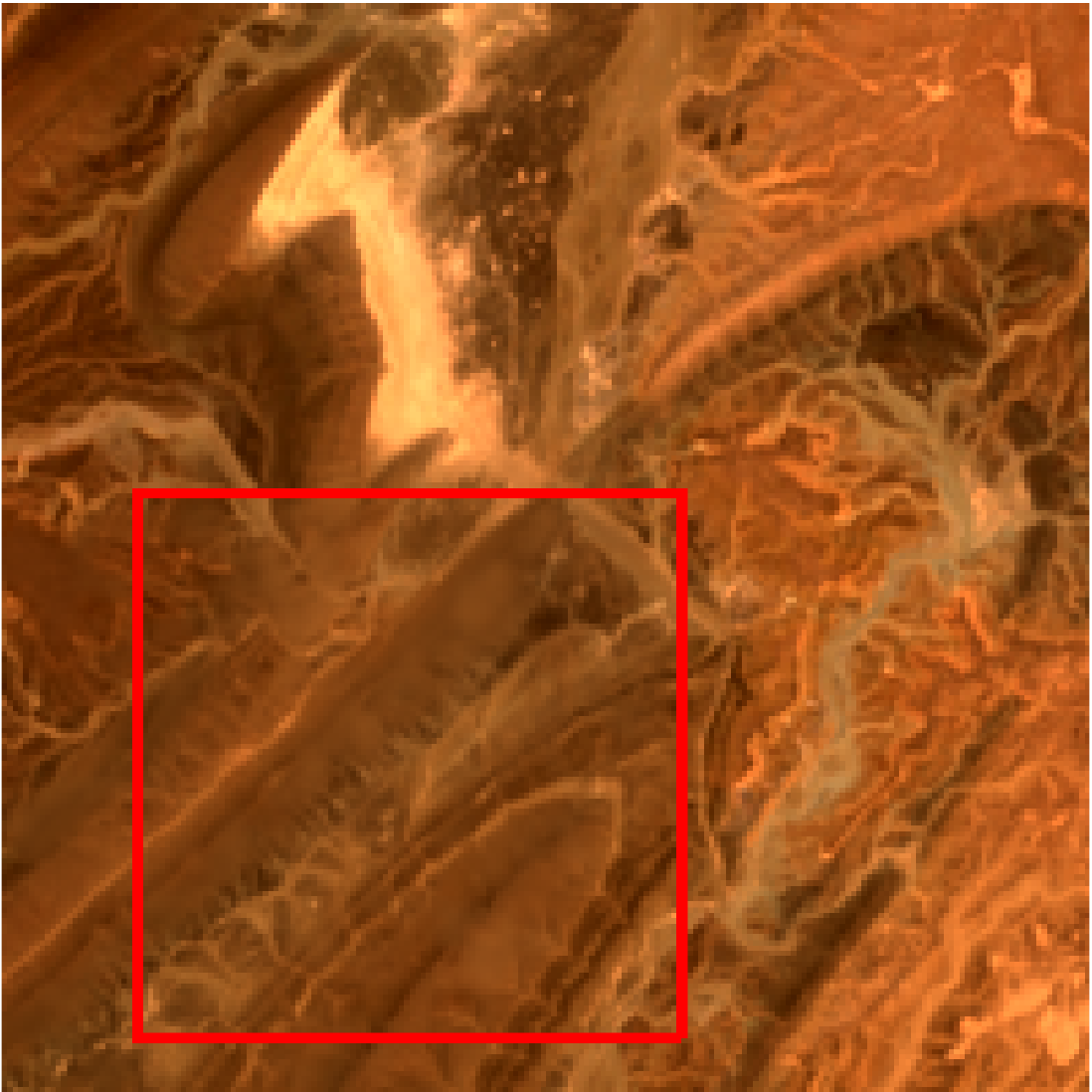}&
			\includegraphics[width=0.12\textwidth]{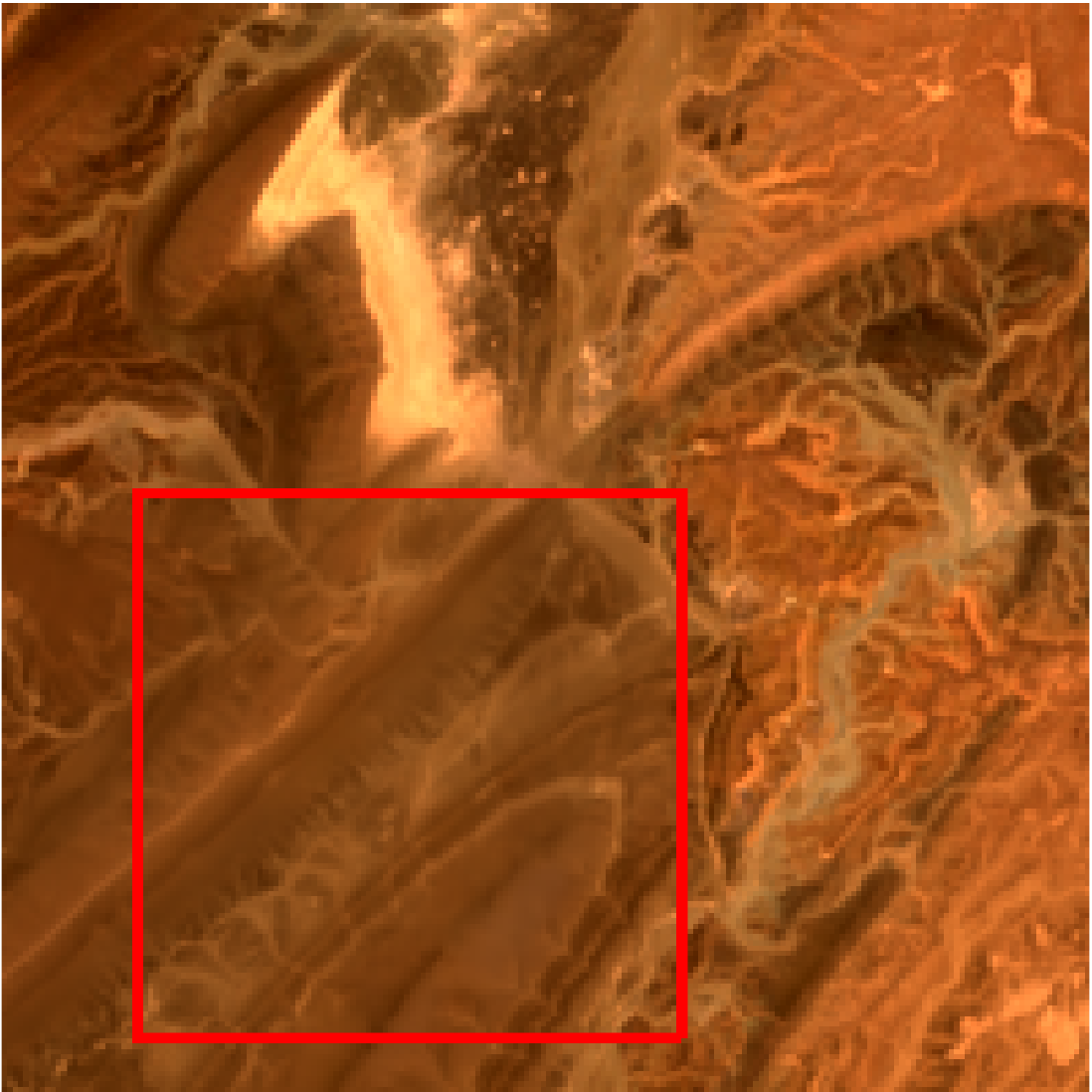}&
			\includegraphics[width=0.12\textwidth]{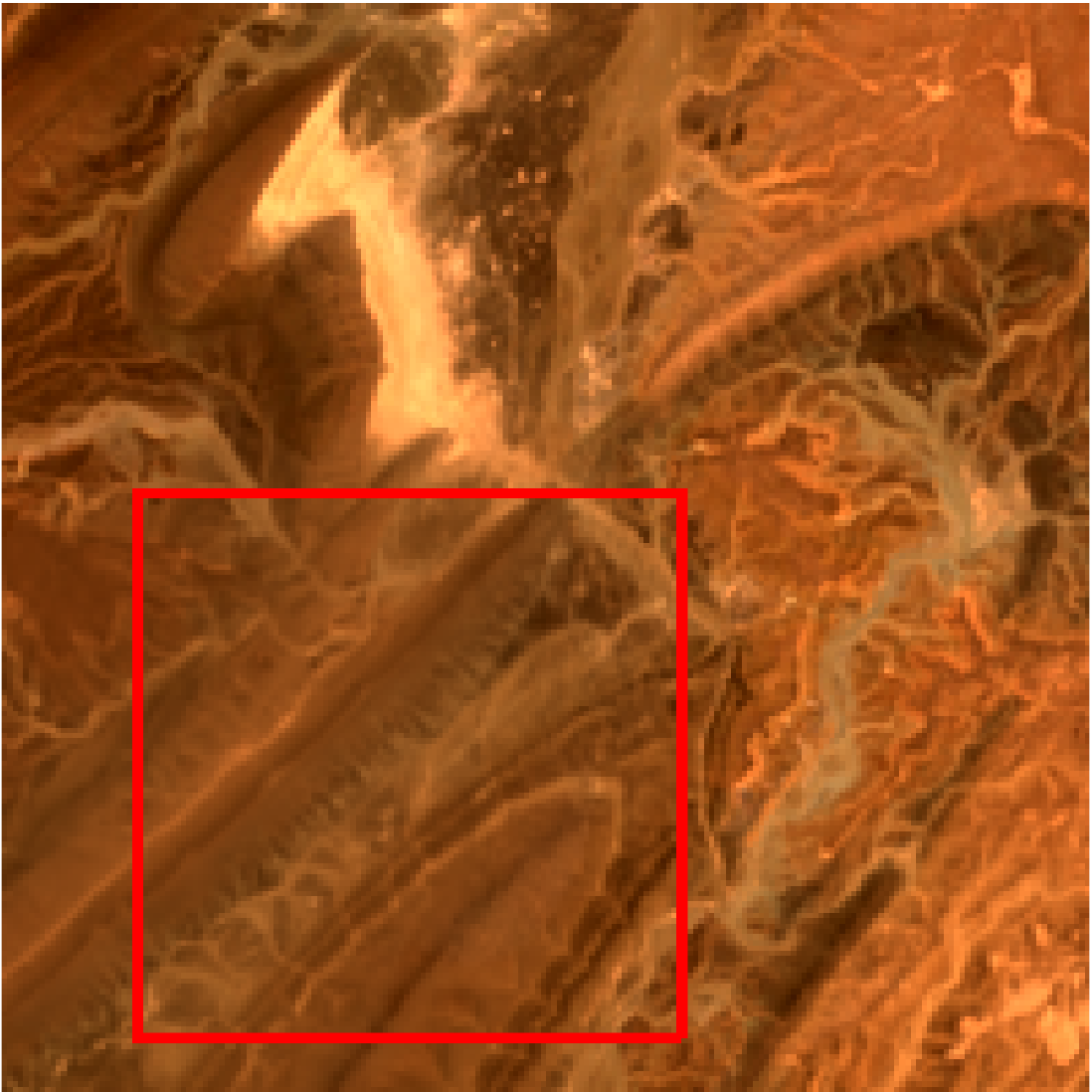}&
			\includegraphics[width=0.12\textwidth]{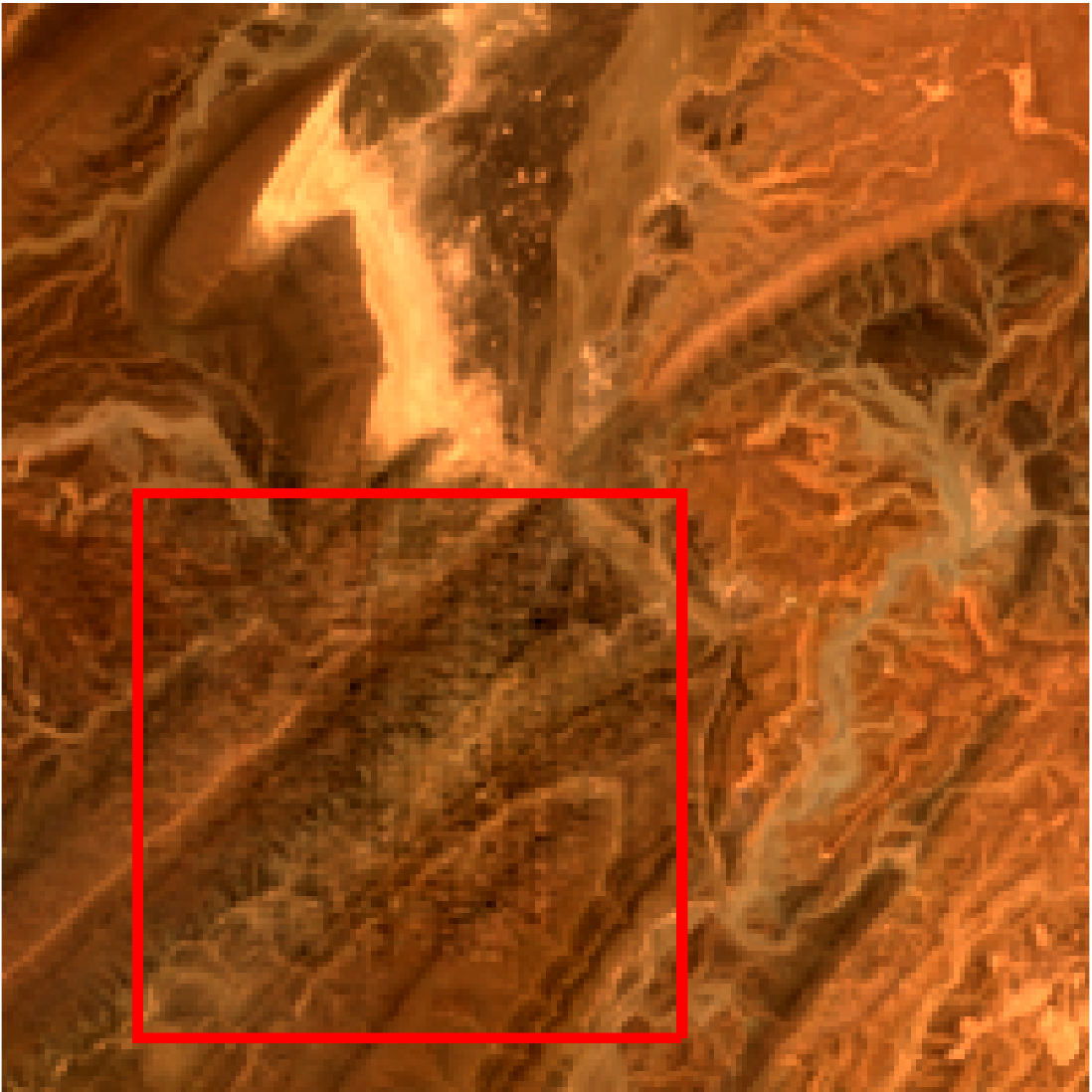}&
			\includegraphics[width=0.12\textwidth]{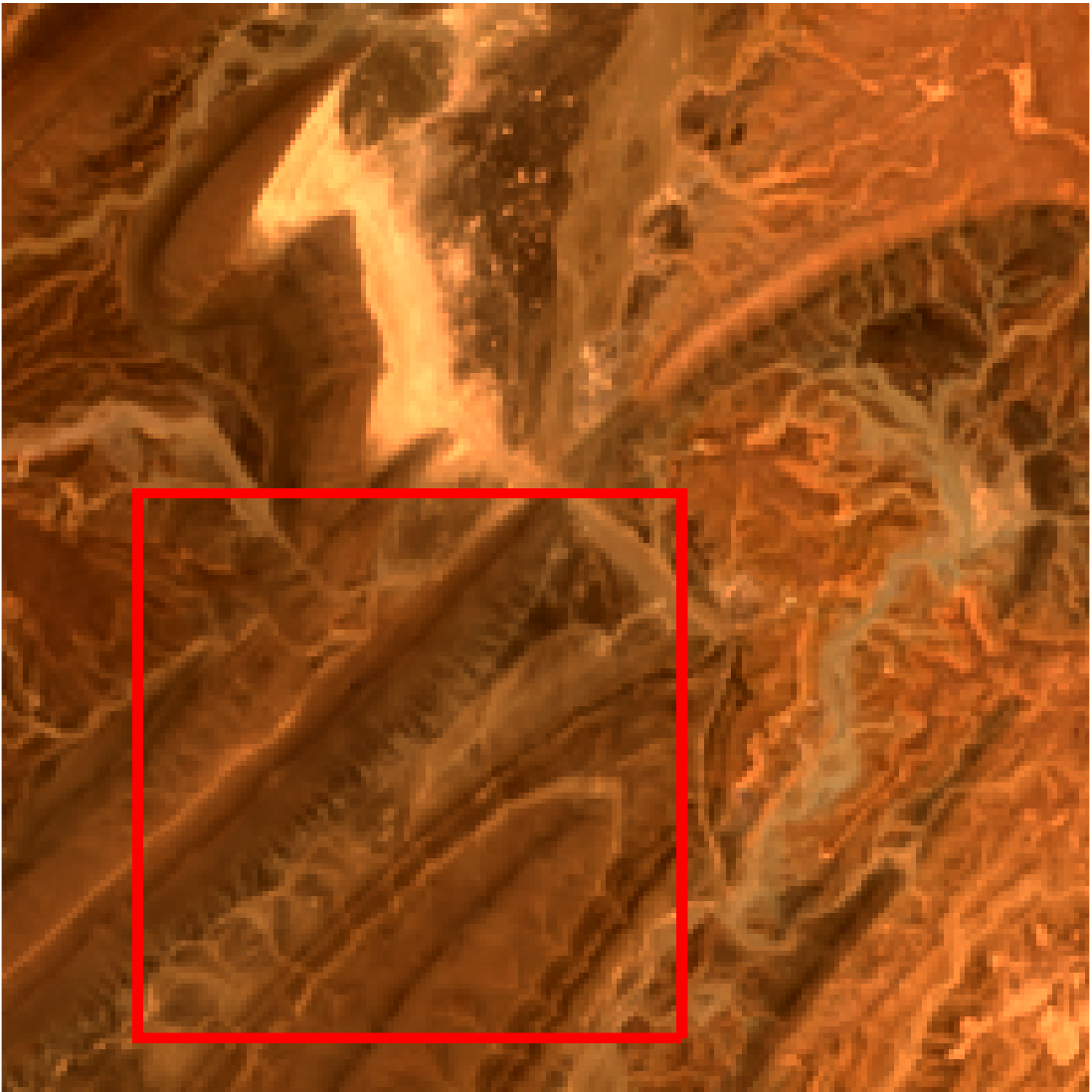}&
			\includegraphics[width=0.12\textwidth]{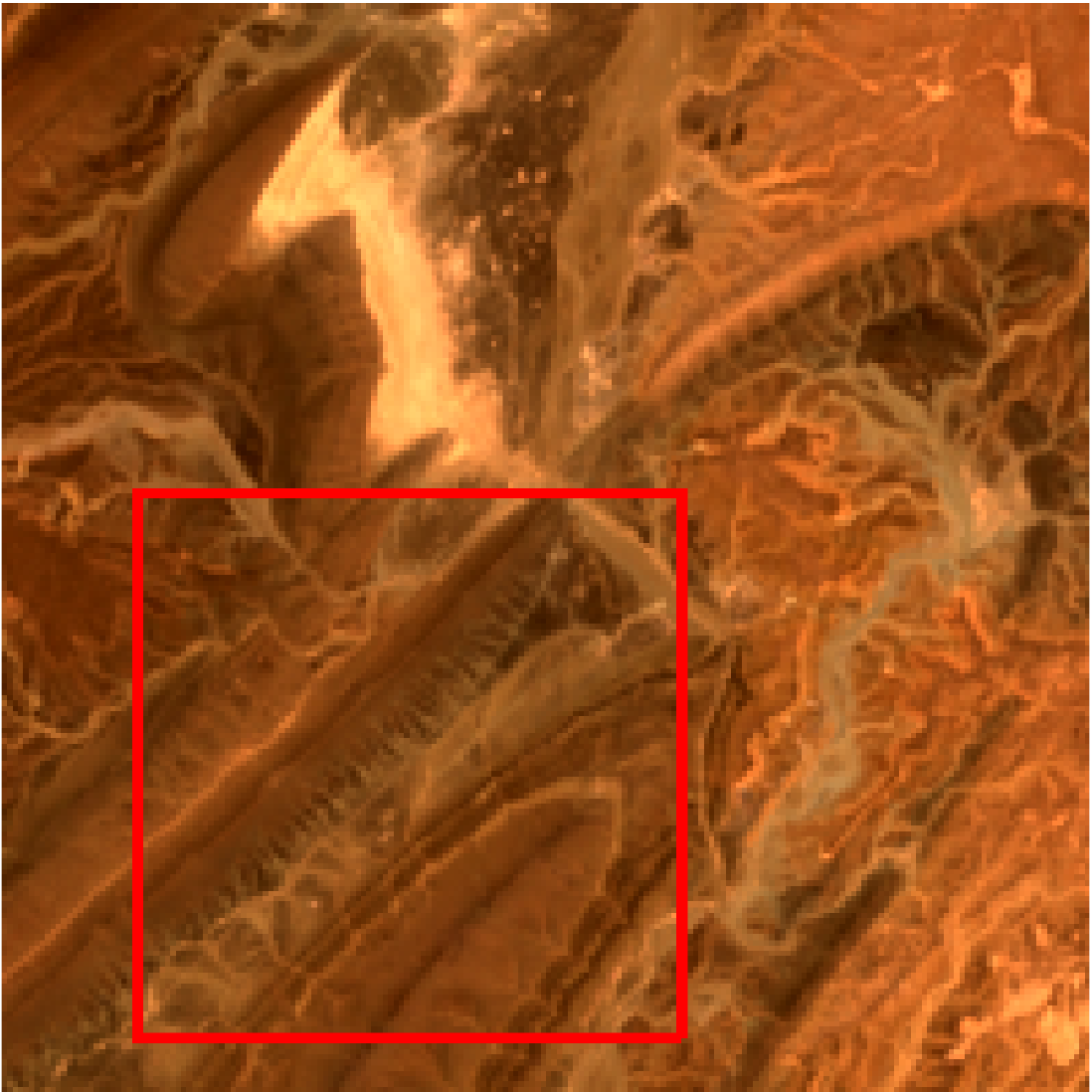}\\
			\includegraphics[width=0.12\textwidth]{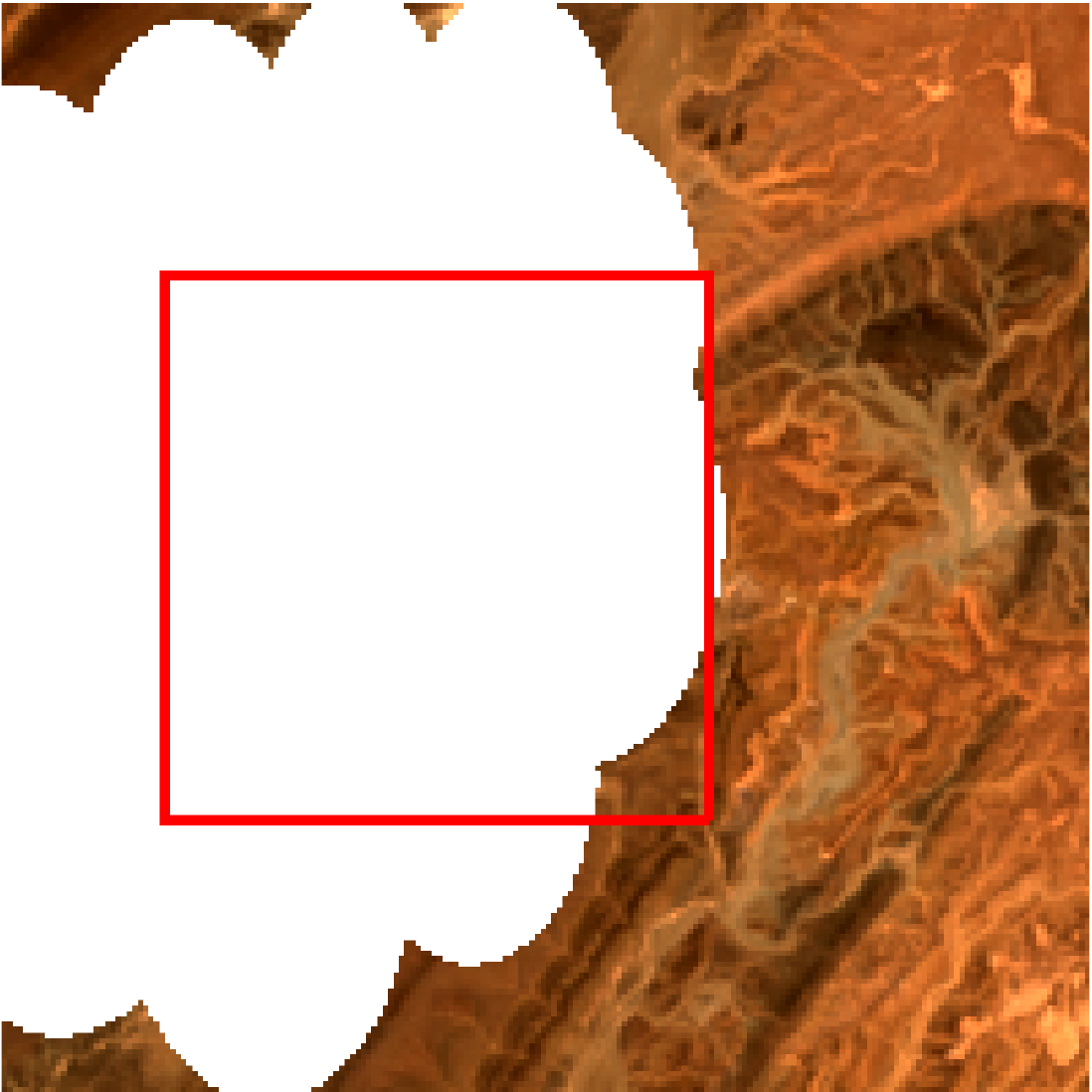}&
			\includegraphics[width=0.12\textwidth]{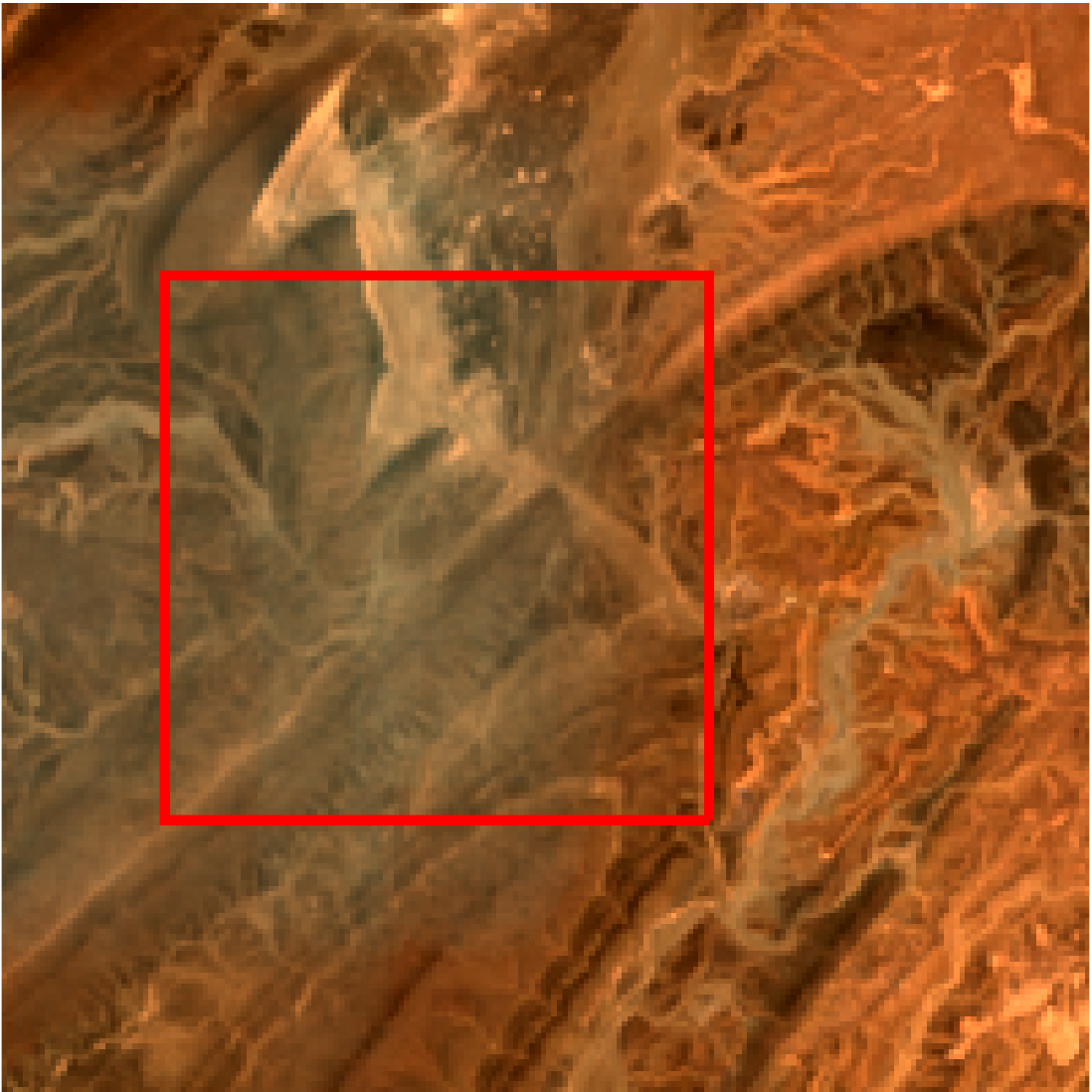}&
			\includegraphics[width=0.12\textwidth]{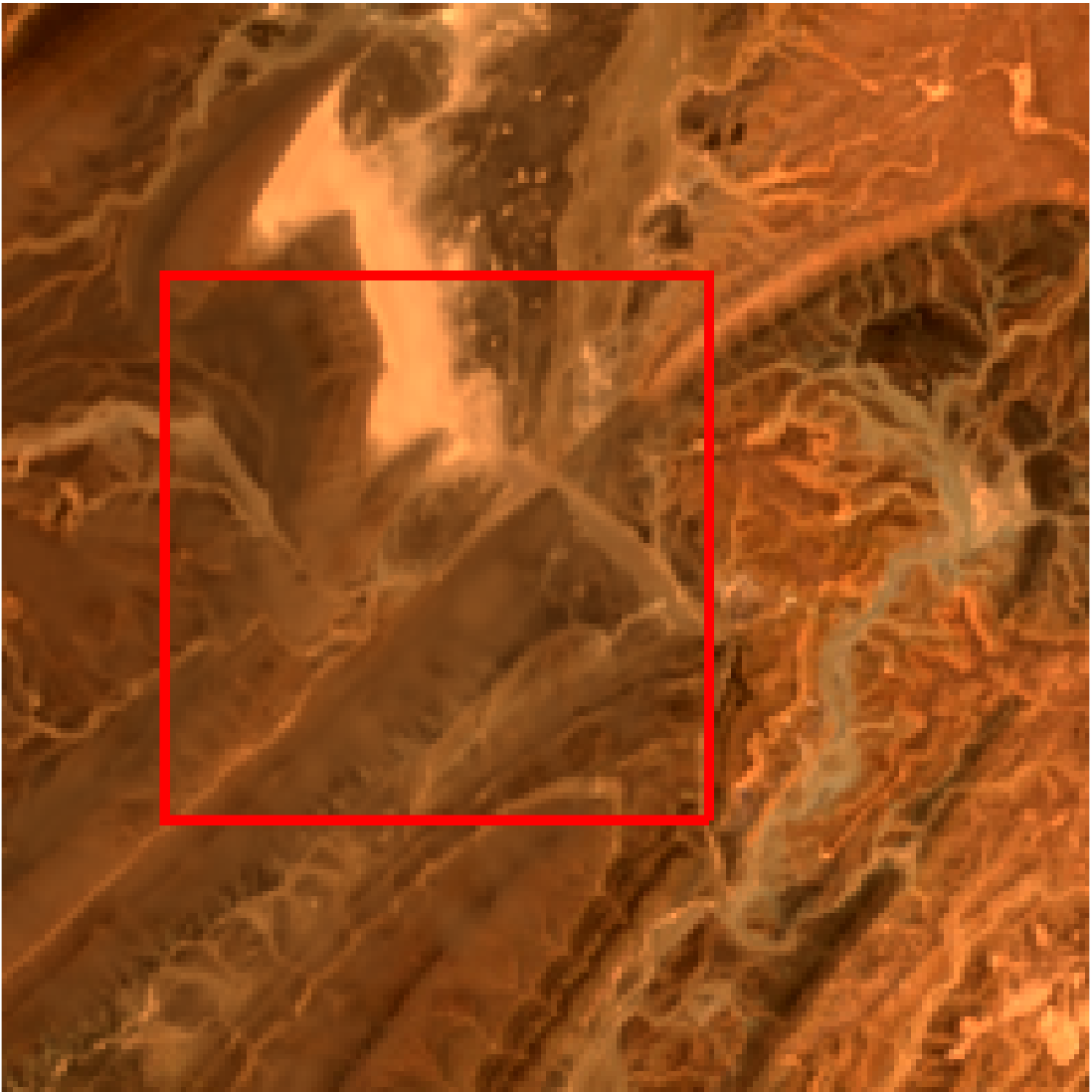}&
			\includegraphics[width=0.12\textwidth]{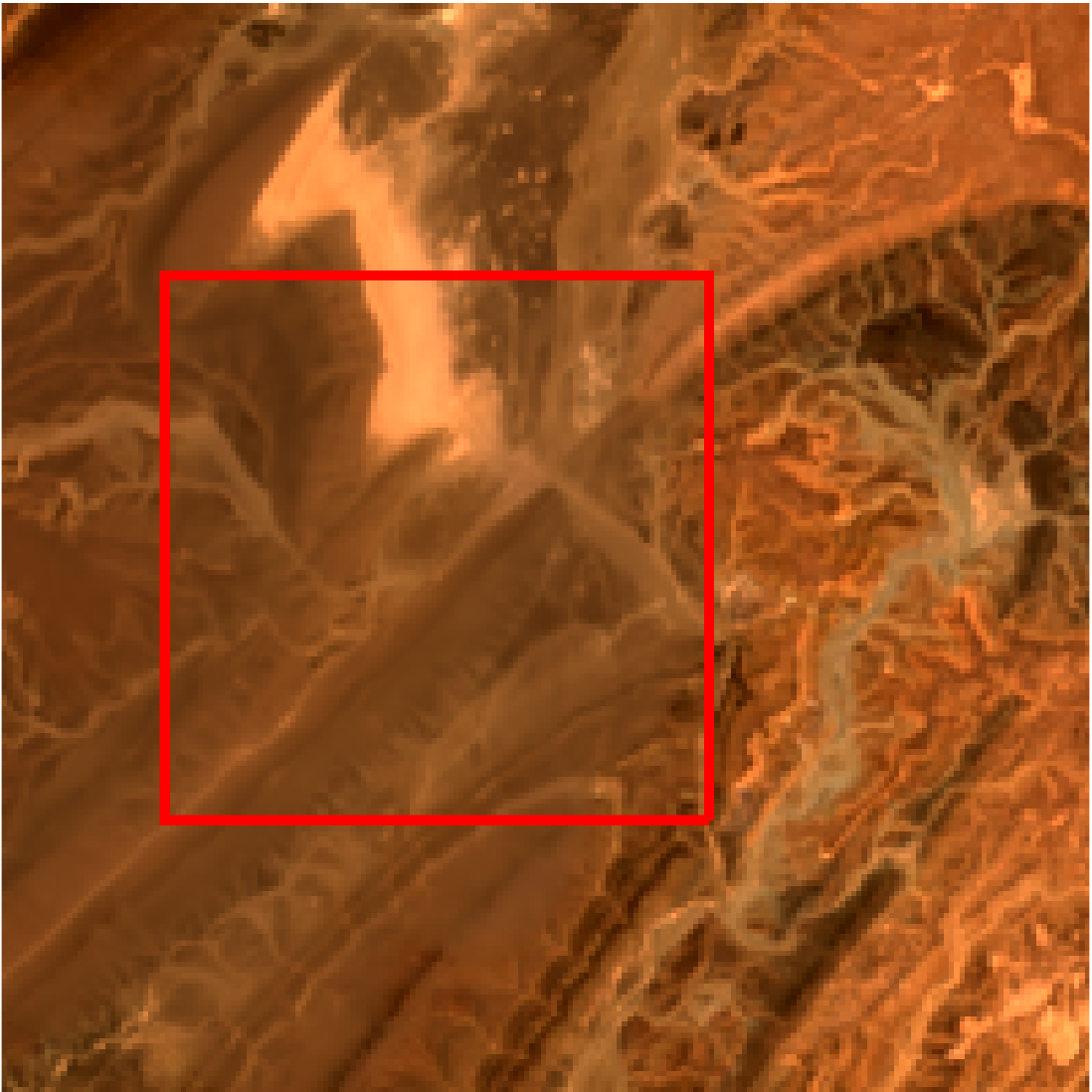}&
			\includegraphics[width=0.12\textwidth]{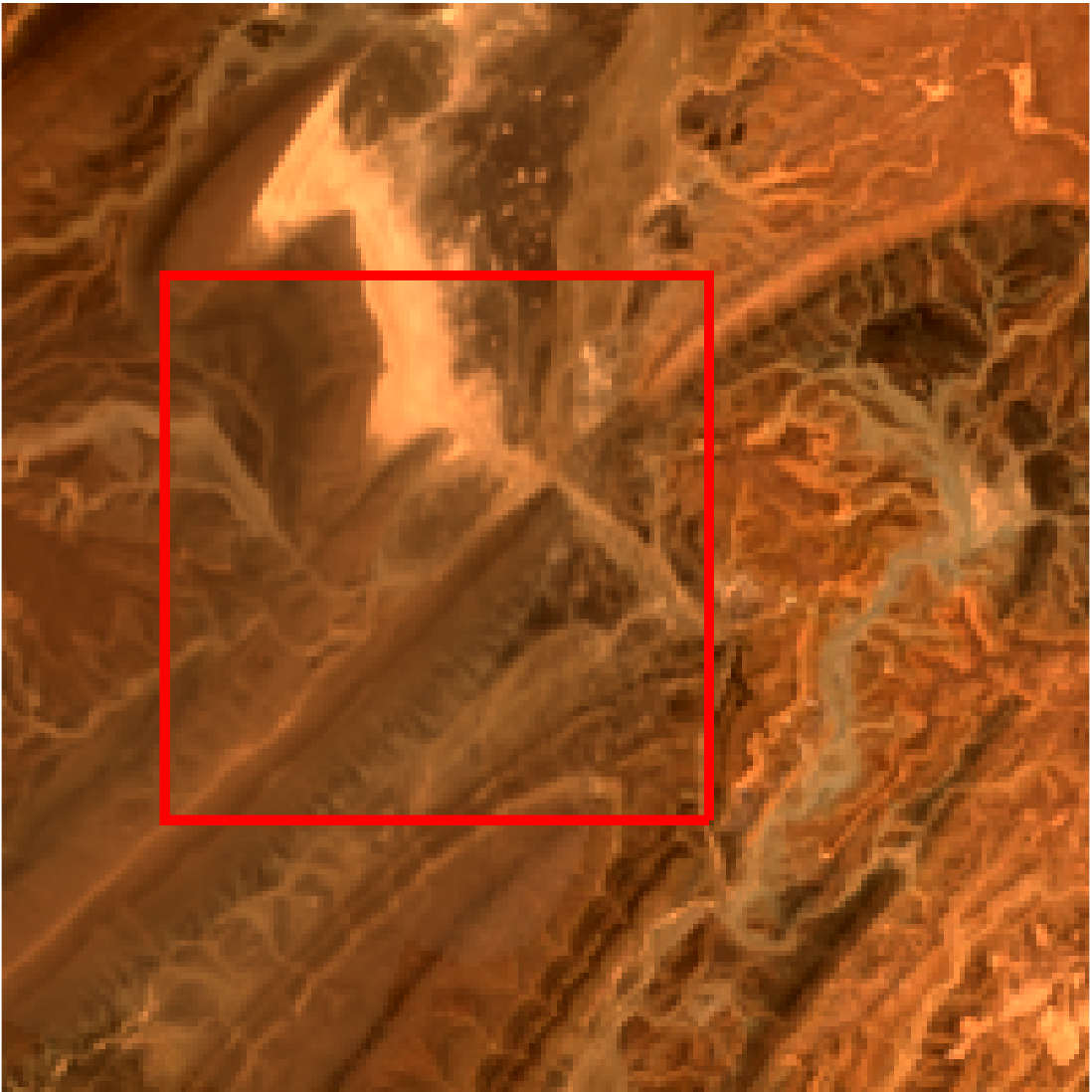}&
			\includegraphics[width=0.12\textwidth]{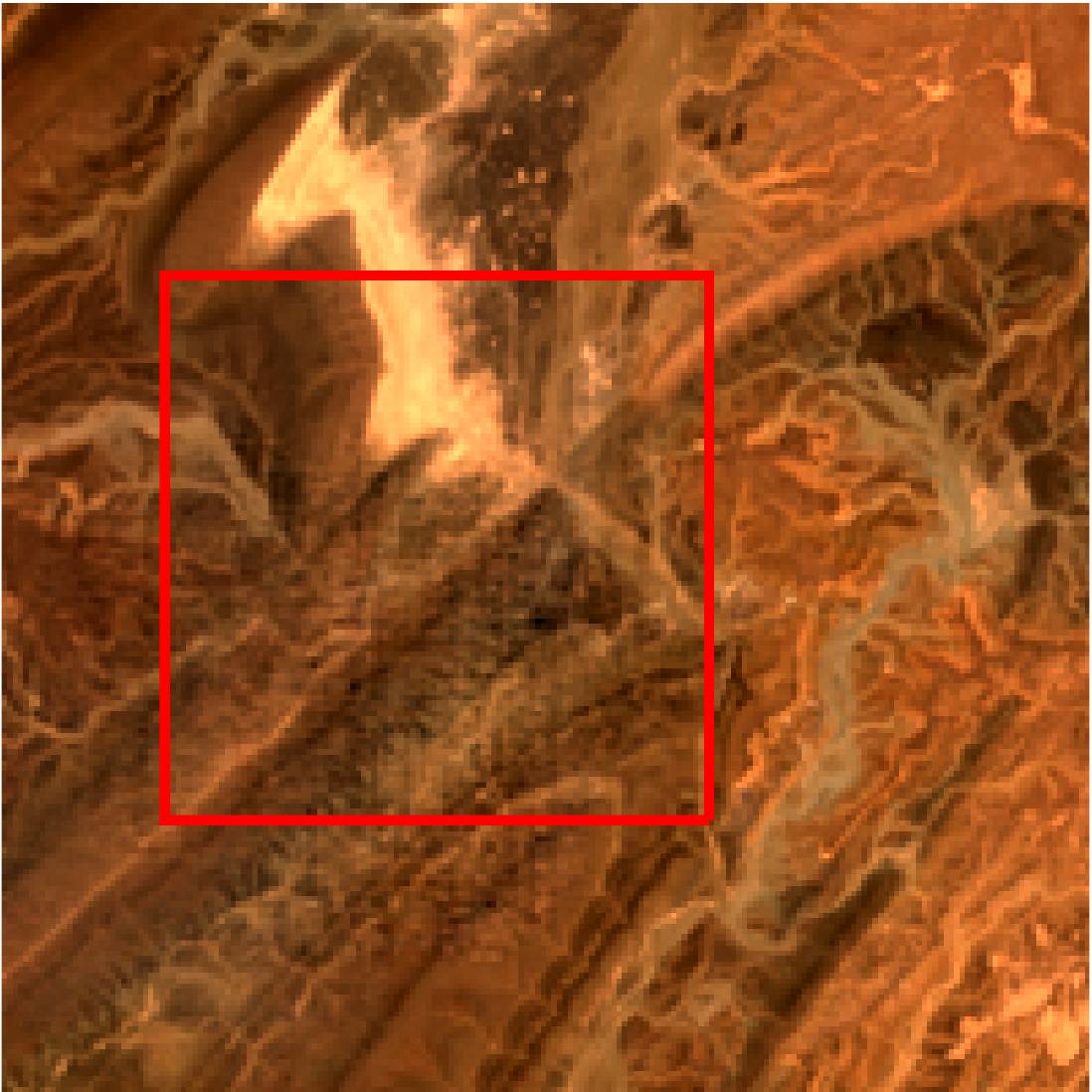}&
			\includegraphics[width=0.12\textwidth]{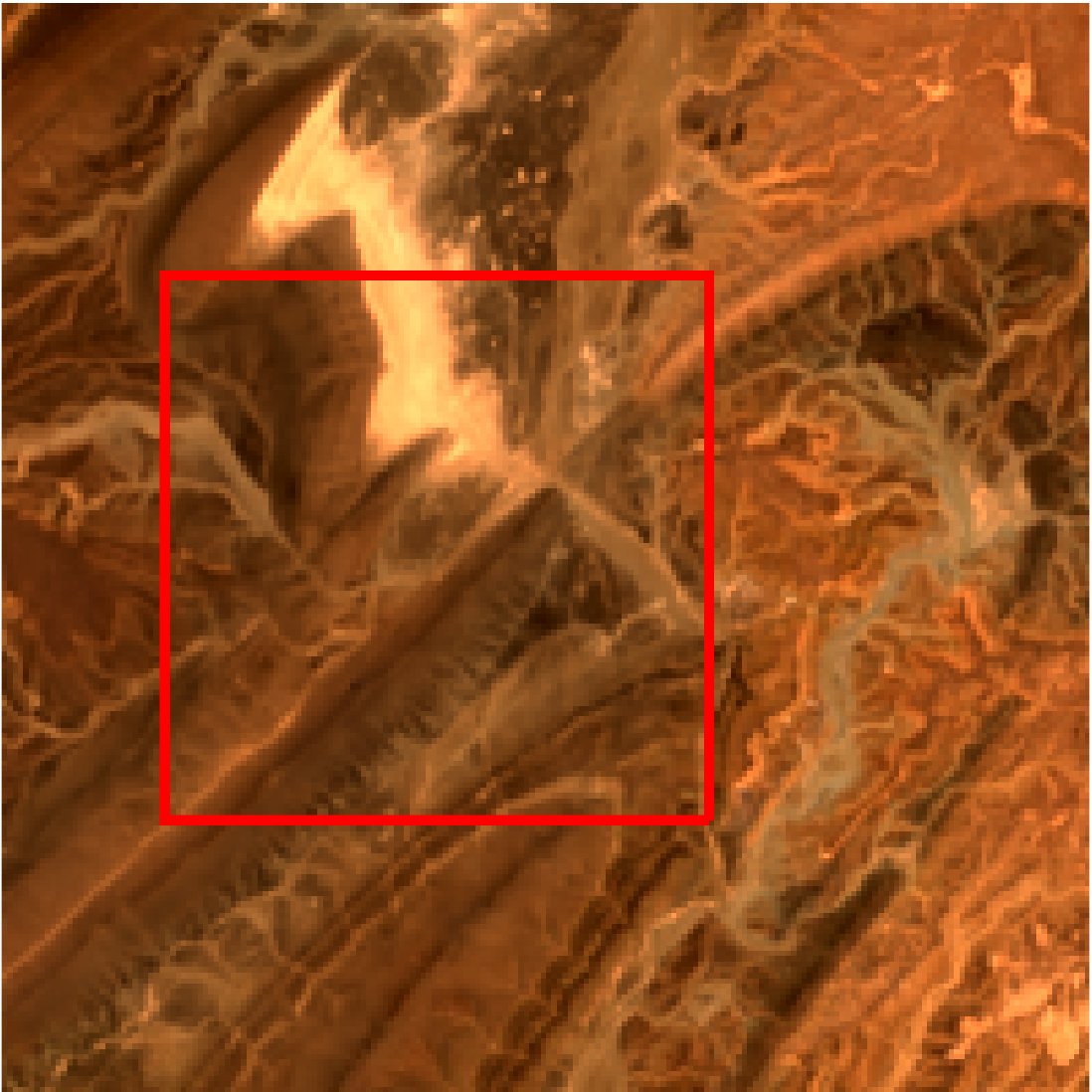}&
			\includegraphics[width=0.12\textwidth]{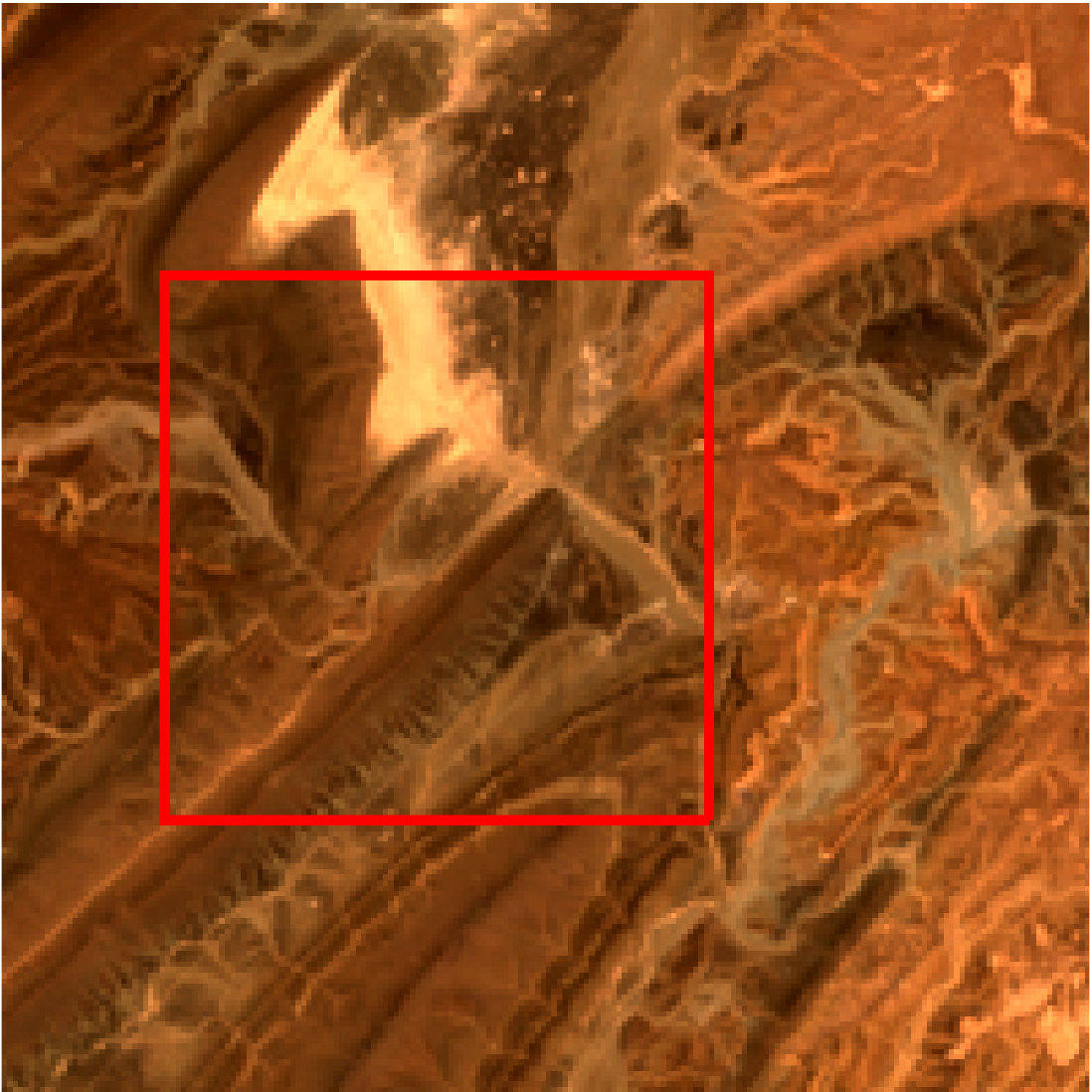}\\
			Degraded image  & KBR-TC  & NL-SNN  & NL-TNN  & NL-TT  & FCTN-TC  & NL-FCTN  & Original image
		\end{tabular}
	\end{center}
	\caption{Cloud removal results of \emph{Morocco}. The pseudo color images are composed of bands 3, 2, and 1. From top to bottom: the results at the second time node and the fourth time node.}\label{mtHSIfig}
	\vspace{-1.5em}
\end{figure*}

\begin{table}[!t]
	\footnotesize
	\setlength{\tabcolsep}{17pt}
	\renewcommand\arraystretch{1.11}
	\caption{The PSNR, SSIM, and SAM values on Morocco of six utilized methods.}
	\begin{center}
		\begin{tabular}{c|ccc}
			\Xhline{0.8pt}
			
			Method    & PSNR    & SSIM    & SAM   \\
			
			\hline
			
			KBR-TC    & 33.8532    & 0.9707    & 2.7453     \\
			NL-SNN    & 46.8914    & 0.9888    & 0.3367     \\
			NL-TNN    & 43.0896    & 0.9810    & 0.4468     \\
			NL-TT     & 39.7771    & 0.9073    & 1.2499     \\
			FCTN-TC   & 44.1284    & 0.9826    & 0.3633     \\
			NL-FCTN   & \textbf{48.1321}    & \textbf{0.9945}    & \textbf{0.2436}     \\ 
			
			\Xhline{0.8pt}
		\end{tabular}
	\end{center}
	\label{mtHSItab}
	\vspace{-1.9em}
\end{table}

\emph{2) Time-Series Sentinel-2 Image Cloud Removal:}  The time-series sentinel-2 image, named \emph{Morocco}\footnotemark[2]\footnotetext[2]{https://theia.cnes.fr/atdistrib/rocket/\#/home.}, is taken over Morocco, which is resized to 200  $\times$ 200 $\times$ 4 $\times$ 6 and rescaled to [0,1]. To simulate the real scene, we assume that all bands of \emph{Morocco} at the same time node are covered by the same type of clouds and at different time nodes are covered by different types of clouds. Table \ref{mtHSItab} reports the PSNR, SSIM, and SAM on cloud removal results of \emph{Morocco} by all compared methods. We observe that the proposed method achieves better results regardless of RSNR, SSIM, or SAM.

Fig. \ref{mtHSIfig} shows the pseudo images of cloud removal results of \emph{Morocco} at two time nodes. Especially in the red box, we observe that KBR-TC can not better preserve the spectral information and causes color loss; NL-SNN, NL-TNN, and NL-TT blur the detail textures of the spatial images; FCTN-TC preserves the details but causes some artifacts; NL-FCTN achieves the closest restoration to the original images among all compared methods.

\section{Conclusion}\label{con}

In this paper, we proposed an NL-FCTN decomposition-based method for RSI inpainting. We introduced FCTN decomposition to the whole RSI and its NSS groups to fully characterize the global correlation and the NSS of RSIs. From another perspective, we combined the FCTN decomposition with the nonlocal patch-based tensor order increment operation, 
which cleverly leverages its remarkable ability on higher-order tensors. The NL-FCTN decomposition-based model was solved by a PAM-based algorithm with a theoretical convergence guarantee. Extensive experiments proved that the proposed method performed better than all compared methods.

\bibliographystyle{ieeetran}
\bibliography{refference}

\end{document}